\documentclass{article}
\usepackage{arxiv}
\usepackage{amsmath,amssymb,amsfonts,amsthm,mathrsfs, rotating}
\usepackage{graphicx,subcaption,booktabs,longtable,multirow}
\usepackage{enumitem}
\usepackage{algorithm}
\usepackage{algpseudocode}
\usepackage{hyperref}
\usepackage{url}

\newcommand{\cf}{\mathrm{cf}}

\theoremstyle{plain}
\newtheorem{thm}{Theorem}[section]
\newtheorem{prop}[thm]{Proposition}
\newtheorem{cor}[thm]{Corollary}
\newtheorem{lemma}[thm]{Lemma}

\theoremstyle{definition}
\newtheorem{defn}[thm]{Definition}
\newtheorem{assn}[thm]{Assumption}

\theoremstyle{remark}
\newtheorem{rem}[thm]{Remark}

\graphicspath{{C:/Users/DELL/Desktop/Papers/Casual_PDE_solver/cf_pde_validation_suite_public_patched_rootfiles/}}

\usepackage{hyperref}
\usepackage{url}
\numberwithin{equation}{section}
\theoremstyle{definition}

\title{Counterfactual Operator Relevance for PDE Discovery: Screening, Pruning, and Identifiability}

\author{ 
Ronald Katende 
Department of Mathematics\\
Kabale University\\
Kikungiri Hill, Katuna Road, 317, Kabale, Uganda\\
\texttt{rkatende92@gmail.com} \\
}

\date{}

\renewcommand{\headeright}{}
\renewcommand{\undertitle}{}
\renewcommand{\shorttitle}{Counterfactual Operator Relevance for PDE Discovery}

\hypersetup{
pdftitle={Counterfactual Operator Relevance for PDE Discovery: Screening, Pruning, and Identifiability},
pdfsubject={q-bio.NC, q-bio.QM},
pdfauthor={R. Katende},
pdfkeywords={PDE discovery, counterfactual operator relevance, sparse residual screening, identifiability, scientific machine learning},
}

\begin{document}
\maketitle

\begin{abstract}
	We study operator relevance in data-driven partial differential equation (PDE) discovery. Sparse residual methods can select terms that improve residual fit, but residual contribution is not the same as functional necessity. We formalize this distinction through counterfactual operator interventions, where a candidate term is deleted or perturbed and the factual and intervened trajectories, or observables, are compared. The resulting theory gives six reusable results. A residual--counterfactual gap theorem shows that deletion effects are governed by the inverse linearized PDE map, not by residual magnitude alone. A certified decision theorem gives error margins for relevance, irrelevance, and abstention under neural or numerical surrogate error. An aliasing theorem characterizes experiment-dependent non-identifiability through the null space of the operator-evaluation design. A constraint-manifold theorem shows that operators vanishing on invariant constraint classes cannot be identified from trajectories restricted to those classes. A pruning-consistency theorem proves that sparse screening followed by counterfactual deletion recovers the functionally relevant support under a recall and margin condition. An observable-level adjoint theorem extends relevance testing from full-state deviations to scientific quantities of interest. Validation experiments test these mechanisms on synthetic PDEs with known support and on public geophysical fields from atmospheric reanalysis and NOAA OISST. The real-data results are reported as operator-surrogate diagnostics, not as unconditional recovery of physical laws. The framework provides a rigorous diagnostic layer for distinguishing residual usefulness from counterfactual operator relevance within a specified library, experiment class, norm, and tolerance.
\end{abstract}

\keywords{PDE discovery \and counterfactual operator relevance \and sparse residual screening \and identifiability \and scientific machine learning} 

\section{Introduction}

Partial differential equations (PDEs) provide compact models for spatially and temporally evolving systems. When the governing operator is known, finite element, spectral, multigrid, operator-splitting, and matrix-free methods provide mature tools for numerical approximation~\cite{brenner2008fem}. In many scientific settings, however, the governing operator is only partially specified. The problem is then not only to solve a PDE, but to infer which candidate terms are needed to explain the observed dynamics.

Physics-informed neural networks (PINNs) impose a prescribed residual during training and are effective when the operator form is known or nearly known~\cite{raissi_physics-informed_2019}. Neural operators and DeepONets learn maps between function spaces and are useful surrogates for forward and inverse problems~\cite{kovachki_neural_2023}. These methods are not, by themselves, operator-discovery procedures. Sparse-regression approaches such as SINDy and PDE-FIND estimate active terms from a candidate library~\cite{brunton_discovering_2016}. Their reliability depends on the library, the conditioning of the induced design matrix, derivative accuracy, and the informativeness of the observed trajectories.

A persistent difficulty is that residual fit is not structural necessity. A term may reduce the residual or receive a nonzero coefficient because it is correlated with other features, because the trajectory excites only a narrow subspace, or because derivative estimates are noisy. Coefficient magnitude and residual contribution can therefore overstate the functional importance of a term.

This paper develops a counterfactual theory of operator relevance for PDE discovery. Given a fitted candidate model, we delete or perturb a candidate operator and compare the induced counterfactual trajectory with the factual trajectory. A term is called relevant only relative to a specified library, experiment class, observation operator, and tolerance. This is a model-based diagnostic of functional necessity. It is not a claim that observational PDE data alone identifies the true physical law.

The paper makes the following contributions. 

We firstly prove a residual-counterfactual gap theorem showing that the effect of deleting an operator is governed by the inverse linearized PDE map, not by residual magnitude alone. We also give a certified counterfactual decision rule. The rule separates relevance, irrelevance, and abstention according to the surrogate error margin.

We characterize operator aliasing through the null space of the finite operator-evaluation design. This identifies when support recovery is impossible from a given experiment. Moreover, we prove constraint-manifold non-identifiability. Operators that vanish on the admissible constraint class cannot be identified from trajectories restricted to that class.

Furthermore, we prove a screening-to-pruning contract. Sparse residual regression is used only to propose a candidate support; counterfactual deletion then recovers the functionally relevant support under a recall and margin condition. Finally, we give an observable-level relevance theorem. It shows how adjoint sensitivities approximate the effect of operator deletion on scientific observables such as mass, energy, flux, or terminal averages.

The paper uses standard sparse-recovery and residual-error estimates only as supporting tools. Lasso screening is treated as a candidate-support mechanism under restricted-eigenvalue or irrepresentability assumptions~\cite{bickel2009simultaneous}. Residual-to-error control is invoked only for fixed well-posed operators under monotonicity-type assumptions~\cite{zeidler1990nonlinear}. The novelty is the counterfactual operator-relevance theory connecting these components.

\subsection{Related Work}

Data-driven discovery of governing equations is commonly formulated as sparse selection from a library of candidate terms. SINDy identifies sparse nonlinear dynamical systems from data, while PDE-FIND extends the idea to PDE terms using derivative estimates and sparse regression~\cite{rudy_data-driven_2017}. Robust variants improve numerical stability through weak formulations, ensembles, neural representations, or alternative regression strategies, but still depend on the identifiability of the induced operator-evaluation design~\cite{fasel2022ensemble}.

Scientific machine learning provides complementary approximation tools. PINNs use automatic differentiation to penalize PDE residuals during training~\cite{raissi_physics-informed_2019}. Neural operators and DeepONets learn solution maps across input functions, parameters, or forcings~\cite{lu_learning_2021}. These methods improve solution approximation, but they do not automatically distinguish generative operator structure from terms that merely improve residual fit.

Causal inference provides a language for interventions and counterfactuals~\cite{pearl_causality_2009}. We use that language in a restricted operator-level sense. An intervention modifies a candidate PDE term, coefficient, source, or constitutive component, and the resulting counterfactual system is solved or approximated. The resulting deviation is a diagnostic for functional operator relevance under the assumed model class. It is weaker than full causal identification, but stronger than ranking terms by coefficient size or residual contribution alone.

\section{Counterfactual PDE Models}
\label{sec:counterfactual-pde-models}

Let \(u:\Omega\times[0,T]\to\mathbb{R}^q\) denote the state of a PDE system and let
\[
\mathcal{N}[u;\gamma]=0
\]
be a candidate governing equation with coefficients, sources, or constitutive parameters collected in \(\gamma\). We use structural causal terminology only to formalize interventions on the fitted operator. The notation does not imply that the full physical causal graph is identifiable from a trajectory.

An intervention replaces part of the operator or its parameters. Given \(\gamma\mapsto\gamma'\), the counterfactual PDE is
\begin{equation}
	\label{eq:cf_pde}
	\mathcal{N}^{\cf}[u^{\cf};\gamma']=0,\qquad
	u^{\cf}(x,0)=u_0^{\cf}(x),\qquad
	\mathcal{B}^{\cf}u^{\cf}=g^{\cf}
	\quad\text{on }\partial\Omega\times(0,T),
\end{equation}
where \(\mathcal{B}^{\cf}\) is the boundary operator. The solution \(u^{\cf}\) is model-based. It is the trajectory implied by the intervened operator, not an independently observed physical outcome unless counterfactual data are available.

The finite counterfactual response is
\begin{equation}
	\label{eq:cf_response}
	\delta(\gamma\to\gamma')
	=
	\|u-u^{\cf}\|_{L^2(\Omega\times(0,T))}.
\end{equation}
Its interpretation is conditional on the operator class, initial and boundary data, observation window, and numerical tolerance.

When \(u^{\cf}\) is not available analytically, we approximate it by a surrogate \(v_\theta\). Let
\[
\mathcal{R}_{\cf}[v_\theta]
=
\mathcal{N}^{\cf}[v_\theta;\gamma']
\]
be the intervened residual. A counterfactual PINN may be trained using
\begin{equation}
	\label{eq:cf_loss}
	\begin{aligned}
		\mathcal{L}_{\cf}(\theta)
		&=
		\|\mathcal{R}_{\cf}[v_\theta]\|^2_{L^2(\Omega\times(0,T))}
		+\lambda_{\mathrm{ic}}\|v_\theta(\cdot,0)-u_0^{\cf}\|^2_{L^2(\Omega)}
		\\
		&\quad
		+\lambda_{\mathrm{bc}}\|\mathcal{B}^{\cf}v_\theta-g^{\cf}\|^2_{L^2(\partial\Omega\times(0,T))}
		+\lambda_{\mathrm{data}}\sum_{i=1}^N |v_\theta(x_i,t_i)-u_i^{\cf}|^2 .
	\end{aligned}
\end{equation}
The last term is used only when counterfactual observations are available. Otherwise, the surrogate is constrained by the intervened residual and by the prescribed initial and boundary data.

Let \(\Gamma=\{\mathcal{T}_1,\ldots,\mathcal{T}_m\}\) be a candidate library. We write the fitted residual as
\begin{equation}
	\label{eq:pde_decomp}
	\mathcal{R}[u;\alpha]
	=
	\mathcal{F}(u)-\sum_{j=1}^m\alpha_j\mathcal{T}_j[u],
\end{equation}
where \(\mathcal{F}\) contains the fixed part of the equation, such as \(\partial_tu\). Let \(u_j^{\cf}\) solve the intervened system obtained by deleting or perturbing \(\mathcal{T}_j\). For a tolerance \(\varepsilon>0\), we call \(\mathcal{T}_j\) counterfactually relevant for the tested experiment if
\begin{equation}
	\label{eq:operator_relevance}
	\delta_j
	=
	\|u-u_j^{\cf}\|_{L^2(\Omega\times(0,T))}
	\ge \varepsilon .
\end{equation}
If \(\delta_j<\varepsilon\), the term is not functionally necessary for that trajectory and intervention. This does not prove physical absence outside the tested setting.

\section{Operator Relevance, Aliasing, and Counterfactual Identifiability}
\label{sec:operator-relevance-theory}

Throughout this section, \(V\) and \(Y\) are Banach spaces, \(u\in V\), and
\begin{equation}
	\label{eq:abstract_residual}
	\mathcal{R}(u;\alpha)
	=
	\mathcal{F}(u)-\sum_{j=1}^m\alpha_j\mathcal{T}_j[u],
	\qquad
	\mathcal{T}_j:V\to Y .
\end{equation}
All claims are relative to the chosen library, experiment class, observation operator, and boundary or initial data.

\subsection{Residual contribution and counterfactual response}

For a fitted coefficient vector \(\alpha\), define the residual contribution
\begin{equation}
	\label{eq:residual_contribution_general}
	C_j(u;\alpha)
	=
	|\alpha_j|\|\mathcal{T}_j[u]\|_Y .
\end{equation}
This is a residual-space quantity. Counterfactual relevance depends on how this direction is propagated by the PDE solution map.

\begin{assn}[Local deletion branch]
	\label{ass:local_solution_branch}
	Let \(\mathcal{R}:V\times\mathbb{R}^m\to Y\) be twice continuously Fréchet differentiable near \((u,\alpha)\), with \(\mathcal{R}(u;\alpha)=0\). For \(j\in[m]\) and \(\rho\in[0,1]\), set
	\[
	\alpha^{(j)}(\rho)=\alpha-\rho\alpha_j e_j .
	\]
	Assume that there is a \(C^2\) branch \(\rho\mapsto u_j(\rho)\in V\) satisfying
	\[
	\mathcal{R}(u_j(\rho);\alpha^{(j)}(\rho))=0,
	\qquad
	u_j(0)=u.
	\]
	Assume also that
	\[
	L_u:=\partial_u\mathcal{R}(u;\alpha):V\to Y
	\]
	is a bounded isomorphism.
\end{assn}

The local branch follows from the Banach-space implicit function theorem when the stated differentiability and invertibility hypotheses hold~\cite{zeidler1986nonlinear}.

\begin{thm}[Residual-counterfactual gap]
	\label{thm:residual_counterfactual_gap}
	Under Assumption~\ref{ass:local_solution_branch},
	\begin{equation}
		\label{eq:first_order_gap}
		u_j'(0)
		=
		-
		L_u^{-1}\!\left[\alpha_j\mathcal{T}_j[u]\right].
	\end{equation}
	If
	\[
	M_j:=\sup_{\rho\in[0,1]}\|u_j''(\rho)\|_V<\infty,
	\]
	then, for every \(\rho\in[0,1]\),
	\begin{equation}
		\label{eq:rho_gap}
		\left|
		\|u_j(\rho)-u\|_V
		-
		\rho\left\|L_u^{-1}\!\left[\alpha_j\mathcal{T}_j[u]\right]\right\|_V
		\right|
		\le
		\frac{\rho^2M_j}{2}.
	\end{equation}
	In particular, residual contribution \(C_j(u;\alpha)\) need not be comparable to counterfactual response unless \(L_u^{-1}\) is controlled on the operator direction \(\alpha_j\mathcal{T}_j[u]\).
\end{thm}

\begin{proof}
	Differentiate
	\[
	\mathcal{R}(u_j(\rho);\alpha-\rho\alpha_j e_j)=0
	\]
	at \(\rho=0\). Since
	\[
	\partial_\alpha\mathcal{R}(u;\alpha)[-\alpha_je_j]
	=
	\alpha_j\mathcal{T}_j[u],
	\]
	we get
	\[
	L_u u_j'(0)+\alpha_j\mathcal{T}_j[u]=0,
	\]
	which proves \eqref{eq:first_order_gap}. Taylor's formula gives
	\[
	u_j(\rho)-u
	=
	\rho u_j'(0)+\int_0^\rho(\rho-s)u_j''(s)\,ds .
	\]
	Thus
	\[
	\left\|u_j(\rho)-u-\rho u_j'(0)\right\|_V
	\le
	\frac{\rho^2M_j}{2}.
	\]
	The result follows from the reverse triangle inequality.
\end{proof}

\begin{cor}[When residual ranking is valid]
	\label{cor:residual_ranking_valid}
	Assume the hypotheses of Theorem~\ref{thm:residual_counterfactual_gap}. Suppose that for some \(0<c_-\le c_+<\infty\),
	\[
	c_-\|g\|_Y
	\le
	\|L_u^{-1}g\|_V
	\le
	c_+\|g\|_Y
	\qquad
	\forall g\in
	\operatorname{span}\{\alpha_j\mathcal{T}_j[u]\}_{j=1}^m .
	\]
	Then
	\[
	\rho c_-C_j(u;\alpha)-\frac{\rho^2M_j}{2}
	\le
	\|u_j(\rho)-u\|_V
	\le
	\rho c_+C_j(u;\alpha)+\frac{\rho^2M_j}{2}.
	\]
\end{cor}

\begin{proof}
	Apply Theorem~\ref{thm:residual_counterfactual_gap} with \(g=\alpha_j\mathcal{T}_j[u]\).
\end{proof}

\subsection{Certified counterfactual decisions}

\begin{thm}[Robust relevance decision]
	\label{thm:robust_cf_decision}
	Let \(u,u_j^{\cf}\in V\) be exact factual and intervened solutions. Let \(u_\theta,v_{\theta,j}\in V\) satisfy
	\[
	\|u-u_\theta\|_V\le\eta_f,
	\qquad
	\|u_j^{\cf}-v_{\theta,j}\|_V\le\eta_{\cf}.
	\]
	Define
	\[
	\delta_j=\|u-u_j^{\cf}\|_V,
	\qquad
	\widehat{\delta}_j=\|u_\theta-v_{\theta,j}\|_V .
	\]
	Then
	\begin{equation}
		\label{eq:cf_decision_stability}
		|\delta_j-\widehat{\delta}_j|
		\le
		\eta_f+\eta_{\cf}.
	\end{equation}
	Consequently,
	\[
	\widehat{\delta}_j>\varepsilon+\eta_f+\eta_{\cf}
	\quad\Longrightarrow\quad
	\delta_j>\varepsilon,
	\]
	and
	\[
	\widehat{\delta}_j<\varepsilon-\eta_f-\eta_{\cf}
	\quad\Longrightarrow\quad
	\delta_j<\varepsilon.
	\]
	If \(|\widehat{\delta}_j-\varepsilon|\le\eta_f+\eta_{\cf}\), the surrogate evidence is inconclusive at tolerance \(\varepsilon\).
\end{thm}

\begin{proof}
	The reverse triangle inequality gives
	\[
	|\delta_j-\widehat{\delta}_j|
	\le
	\|(u-u_j^{\cf})-(u_\theta-v_{\theta,j})\|_V
	\le
	\|u-u_\theta\|_V+\|u_j^{\cf}-v_{\theta,j}\|_V .
	\]
	The threshold implications follow directly.
\end{proof}

\begin{cor}[Residual-certified decision]
	\label{cor:residual_certified_decision}
	If residual-to-error bounds
	\[
	\|u-u_\theta\|_V\le a_f\|A_f(u_\theta)\|_{Y_f},
	\qquad
	\|u_j^{\cf}-v_{\theta,j}\|_V\le a_{\cf}\|A_{\cf}(v_{\theta,j})\|_{Y_{\cf}}
	\]
	hold for the factual and counterfactual operators, then Theorem~\ref{thm:robust_cf_decision} applies with
	\[
	\eta_f=a_f\|A_f(u_\theta)\|_{Y_f},
	\qquad
	\eta_{\cf}=a_{\cf}\|A_{\cf}(v_{\theta,j})\|_{Y_{\cf}} .
	\]
\end{cor}

\begin{proof}
	Substitute the stated estimates into Theorem~\ref{thm:robust_cf_decision}.
\end{proof}

\subsection{Operator aliasing}

Let \(\mathcal{E}\) be a finite family of experiments, with trajectories \(u^{(e)}\) and sampling points \(\{z_i^{(e)}\}_{i=1}^{n_e}\). Define
\[
A_{\mathcal{E}}\alpha
=
\left(
\sum_{j=1}^m\alpha_j\mathcal{T}_j[u^{(e)}](z_i^{(e)})
\right)_{e,i}.
\]

\begin{defn}[Experimental aliasing]
	\label{def:experimental_aliasing}
	Two vectors \(\alpha,\tilde{\alpha}\in\mathbb{R}^m\) are aliased on \(\mathcal{E}\) if
	\[
	A_{\mathcal{E}}(\alpha-\tilde{\alpha})=0.
	\]
	The finite-design identifiable object is
	\[
	[\alpha]_{\mathcal{E}}
	=
	\{\tilde{\alpha}:A_{\mathcal{E}}\tilde{\alpha}=A_{\mathcal{E}}\alpha\}.
	\]
\end{defn}

\begin{thm}[Aliasing obstruction]
	\label{thm:aliasing_obstruction}
	If \(\alpha\neq\tilde{\alpha}\) are aliased on \(\mathcal{E}\), then no residual-regression procedure depending on the experiment only through \(A_{\mathcal{E}}\alpha\) can distinguish \(\alpha\) from \(\tilde{\alpha}\). If their supports differ, exact support recovery is impossible on \(\mathcal{E}\).
\end{thm}

\begin{proof}
	Aliased vectors give the same input to any such procedure:
	\[
	A_{\mathcal{E}}\alpha=A_{\mathcal{E}}\tilde{\alpha}.
	\]
	The output must therefore be identical for both. If the supports differ, the common output cannot equal both supports.
\end{proof}

\begin{cor}[Single-mode reaction-diffusion aliasing]
	\label{cor:single_mode_aliasing}
	Let \(-\Delta\varphi_k=\lambda_k\varphi_k\), and suppose
	\[
	u(x,t)=a(t)\varphi_k(x).
	\]
	Then
	\[
	\Delta u=-\lambda_k u.
	\]
	Hence, in
	\[
	\partial_tu=\kappa u+D\Delta u,
	\]
	the experiment identifies only
	\[
	\kappa-\lambda_kD,
	\]
	not \(\kappa\) and \(D\) separately.
\end{cor}

\begin{proof}
	Since \(\Delta u=a(t)\Delta\varphi_k=-\lambda_k u\),
	\[
	\kappa u+D\Delta u=(\kappa-\lambda_kD)u.
	\]
\end{proof}

\subsection{Separability from multiple experiments}

Let \(A_K\) be the stacked operator-evaluation matrix from \(K\) experiments. For a support \(S\), define
\[
\widehat{\kappa}(A_K;S)
=
\lambda_{\min}\!\left(\frac{1}{n_K}A_{K,S}^{\top}A_{K,S}\right).
\]

\begin{thm}[Multi-experiment separability]
	\label{thm:multi_experiment_separability}
	Let \(A_K\) be the stacked operator-evaluation matrix from \(K\) experiments, and let \(S\subset[m]\) be a target support. If \(A_{K,S}\) has full column rank, then the active columns indexed by \(S\) are linearly distinguishable on the stacked experiment. If, in addition, the anchored regression model satisfies the usual noise, beta-min, and restricted-eigenvalue or irrepresentability conditions on \(A_K\), then the corresponding sparse-regression screening or signed-support guarantees apply.
\end{thm}

\begin{proof}
	Full column rank of \(A_{K,S}\) is equivalent to positive definiteness of \(A_{K,S}^{\top}A_{K,S}/n_K\). This excludes aliasing among columns in \(S\). The final claim follows from the standard Lasso screening and sign-consistency results under restricted-eigenvalue or irrepresentability assumptions, together with their usual noise and beta-min conditions~\cite{bickel2009simultaneous,zhao2006model}.
\end{proof}

\begin{prop}[Population spectrum under added experiments]
	\label{prop:population_spectrum_added_experiments}
	Let
	\[
	\Sigma_K=\sum_{k=1}^K w_k\Sigma^{(k)},\qquad w_k>0,
	\]
	where each \(\Sigma^{(k)}\) is positive semidefinite. If \(\Sigma^{(K+1)}\succeq0\), then
	\[
	\Sigma_{K+1}
	=
	\Sigma_K+w_{K+1}\Sigma^{(K+1)}
	\succeq
	\Sigma_K .
	\]
	For every fixed support \(S\),
	\[
	\lambda_{\min}\big((\Sigma_{K+1})_{SS}\big)
	\ge
	\lambda_{\min}\big((\Sigma_K)_{SS}\big).
	\]
\end{prop}

\begin{proof}
	For every vector \(v\),
	\[
	v^\top(\Sigma_{K+1}-\Sigma_K)v
	=
	w_{K+1}v^\top\Sigma^{(K+1)}v
	\ge0.
	\]
	The eigenvalue inequality follows from the Rayleigh quotient.
\end{proof}

\subsection{Constraint-manifold non-identifiability}

\begin{thm}[Constraint-null non-identifiability]
	\label{thm:constraint_null_nonidentifiability}
	Let \(M\subset V\) and let \(\mathcal{T}:V\to Y\) satisfy
	\[
	\mathcal{T}[u]=0
	\qquad
	\forall u\in M.
	\]
	If all observed trajectories lie in \(M\), then the coefficient of \(\mathcal{T}\) is not identifiable from residual evaluations on those trajectories. More precisely, for every \(a\in\mathbb{R}\),
	\[
	\mathcal{R}(u^{(e)};\alpha+a e_{\mathcal{T}})
	=
	\mathcal{R}(u^{(e)};\alpha)
	\]
	for every observed trajectory \(u^{(e)}\).
\end{thm}

\begin{proof}
	Since \(u^{(e)}\in M\), \(\mathcal{T}[u^{(e)}]=0\). Adding \(a\mathcal{T}\) changes no evaluated residual.
\end{proof}

\begin{cor}[Maxwell divergence constraint]
	\label{cor:maxwell_constraint}
	For Maxwell trajectories satisfying \(\nabla\cdot E=0\),
	\[
	\nabla(\nabla\cdot E)=0.
	\]
	Thus the coefficient of \(\nabla(\nabla\cdot E)\) is not identifiable from data restricted to divergence-free electric fields.
\end{cor}

\begin{cor}[Leray-projected incompressible flow]
	\label{cor:leray_pressure}
	Let \(P\) be the Leray projection onto divergence-free vector fields. Since
	\[
	P\nabla p=0,
	\]
	pressure is not identifiable as an independent sparse term in the projected velocity equation, although it remains part of the constrained Navier--Stokes formulation~\cite{temam1977navier}.
\end{cor}

\subsection{Consistency of screening followed by counterfactual pruning}

Let
\[
S_{\cf}^{\star}
=
\{j:\delta_j\ge\varepsilon\}
\]
be the exact counterfactual relevance set at tolerance \(\varepsilon\).

\begin{thm}[Counterfactual pruning consistency]
	\label{thm:cf_pruning_consistency}
	Assume that sparse screening returns \(S_{\mathrm{fit}}\) satisfying
	\[
	S_{\cf}^{\star}\subseteq S_{\mathrm{fit}}.
	\]
	For each \(j\in S_{\mathrm{fit}}\), suppose
	\[
	|\widehat{\delta}_j-\delta_j|\le\eta .
	\]
	If
	\[
	|\delta_j-\varepsilon|>\eta
	\qquad
	\forall j\in S_{\mathrm{fit}},
	\]
	then
	\[
	\widehat{S}_{\cf}
	=
	\{j\in S_{\mathrm{fit}}:\widehat{\delta}_j\ge\varepsilon\}
	\]
	satisfies
	\[
	\widehat{S}_{\cf}=S_{\cf}^{\star}.
	\]
\end{thm}

\begin{proof}
	If \(j\in S_{\cf}^{\star}\), then \(\delta_j\ge\varepsilon\). The margin gives \(\delta_j>\varepsilon+\eta\), hence \(\widehat{\delta}_j>\varepsilon\). If \(j\notin S_{\cf}^{\star}\), then \(\delta_j<\varepsilon\). The margin gives \(\delta_j<\varepsilon-\eta\), hence \(\widehat{\delta}_j<\varepsilon\). Since \(S_{\cf}^{\star}\subseteq S_{\mathrm{fit}}\), equality follows.
\end{proof}

\section{Sparse Residual Screening and Reusable Design Conditions}
\label{sec:sparse-screening-design}

Sparse regression is used here as a screening step. It proposes a finite candidate support from an operator-evaluation design matrix. It does not, by itself, establish counterfactual relevance. The role of this section is to state the finite-sample conditions under which sparse screening is reliable enough to feed the counterfactual deletion step.

\subsection{Anchored operator regression}

A homogeneous PDE relation
\begin{equation}
	\label{eq:homogeneous_operator_relation}
	\sum_{j=0}^m \alpha_j^\star \mathcal{T}_j[u]=0,
	\qquad
	\alpha^\star\neq0,
\end{equation}
does not identify the global scale of \(\alpha^\star\). We fix scale by choosing an anchor \(\mathcal{T}_0\) with \(\alpha_0^\star\neq0\), and define
\[
\beta_j^\star=-\frac{\alpha_j^\star}{\alpha_0^\star},
\qquad j=1,\ldots,m .
\]
Then
\begin{equation}
	\label{eq:anchored_regression_identity}
	\mathcal{T}_0[u]
	=
	\sum_{j=1}^m \beta_j^\star \mathcal{T}_j[u].
\end{equation}
Given sampling points \(z_i=(x_i,t_i)\), define
\begin{equation}
	\label{eq:operator_design}
	y_i=\mathcal{T}_0[u](z_i),
	\qquad
	A_{ij}=\mathcal{T}_j[u](z_i),
\end{equation}
so that
\begin{equation}
	\label{eq:anchored_regression}
	y=A\beta^\star+\eta .
\end{equation}
Here \(\eta\) includes observation noise, derivative error, solver error, and model mismatch. The support recovered from \(\beta^\star\) determines the support of \(\alpha^\star\) away from the anchor.

\begin{rem}[Why anchoring is necessary]
	Without an anchor, the penalized homogeneous problem
	\[
	\min_{\alpha}\|A\alpha\|_2^2+\lambda\|\alpha\|_1
	\]
	has the trivial minimizer \(\alpha=0\). Thus sparse recovery must be formulated either through anchoring, a scale constraint, or a normalized constrained problem. We use the anchored form \eqref{eq:anchored_regression}.
\end{rem}

\subsection{Restricted-eigenvalue screening}

Let \(S=\operatorname{supp}(\beta^\star)\), \(|S|=s\), and normalize the columns of \(A\) so that
\[
\|A_{\cdot j}\|_2^2/n=1.
\]

\begin{assn}[Restricted eigenvalue]
	\label{ass:RE}
	The matrix \(A\) satisfies a restricted-eigenvalue condition on \(S\) with constant \(\kappa>0\) if
	\[
	\frac{1}{n}\|A\Delta\|_2^2
	\ge
	\kappa\|\Delta\|_2^2
	\]
	for every \(\Delta\in\mathbb{R}^m\) satisfying
	\[
	\|\Delta_{S^c}\|_1\le3\|\Delta_S\|_1 .
	\]
\end{assn}

Consider the Lasso estimator
\begin{equation}
	\label{eq:lasso_screening}
	\hat{\beta}
	\in
	\arg\min_{\beta\in\mathbb{R}^m}
	\left\{
	\frac{1}{2n}\|A\beta-y\|_2^2+\lambda\|\beta\|_1
	\right\}.
\end{equation}

\begin{thm}[Lasso screening error under RE]
	\label{thm:lasso_screening_error}
	Assume \eqref{eq:anchored_regression}, Assumption~\ref{ass:RE}, and
	\[
	\left\|\frac{1}{n}A^\top\eta\right\|_\infty
	\le
	\frac{\lambda}{2}.
	\]
	Then every solution of \eqref{eq:lasso_screening} satisfies
	\begin{equation}
		\label{eq:lasso_error_bounds}
		\|\hat{\beta}-\beta^\star\|_2
		\le
		B_s,
		\qquad
		\|\hat{\beta}-\beta^\star\|_1
		\le
		4\sqrt{s}B_s,
		\qquad
		B_s:=\frac{4\lambda\sqrt{s}}{\kappa}.
	\end{equation}
\end{thm}

\begin{proof}
	This is the standard Lasso oracle inequality under the restricted-eigenvalue condition~\cite{bickel2009simultaneous}. The constants follow from the basic inequality, the cone condition
	\[
	\|\Delta_{S^c}\|_1\le3\|\Delta_S\|_1,
	\]
	and Assumption~\ref{ass:RE}, with \(\Delta=\hat{\beta}-\beta^\star\).
\end{proof}

The theorem has two different uses. A high threshold gives no false positives after screening. A lower threshold gives high recall, which is preferable when counterfactual pruning will remove false positives.

\begin{cor}[Thresholded support control]
	\label{cor:thresholded_support_control}
	Let
	\[
	\widehat{S}_{\tau}
	=
	\{j:|\hat{\beta}_j|>\tau\}.
	\]
	Under Theorem~\ref{thm:lasso_screening_error}:
	
	\begin{enumerate}
		\item If \(\tau\ge B_s\), then
		\[
		\widehat{S}_{\tau}\subseteq S .
		\]
		
		\item If
		\[
		\min_{j\in S}|\beta_j^\star|>\tau+B_s,
		\]
		then
		\[
		S\subseteq \widehat{S}_{\tau}.
		\]
		
		\item If both conditions hold, then
		\[
		\widehat{S}_{\tau}=S .
		\]
	\end{enumerate}
\end{cor}

\begin{proof}
	For \(j\notin S\), \(\beta_j^\star=0\), hence
	\[
	|\hat{\beta}_j|\le\|\hat{\beta}-\beta^\star\|_2\le B_s.
	\]
	Thus \(\tau\ge B_s\) gives \(\widehat{S}_\tau\subseteq S\). For \(j\in S\),
	\[
	|\hat{\beta}_j|
	\ge
	|\beta_j^\star|-|\hat{\beta}_j-\beta_j^\star|
	\ge
	|\beta_j^\star|-B_s.
	\]
	If \(\min_{j\in S}|\beta_j^\star|>\tau+B_s\), then all active indices exceed threshold. The equality statement follows by combining the two inclusions.
\end{proof}

\begin{rem}[Screening for counterfactual pruning]
	The counterfactual pipeline usually needs recall more than exact sparse support. If \(\tau\) is chosen so that \(S\subseteq\widehat{S}_\tau\), false positives can be removed by counterfactual deletion. This is safer than using a high threshold that may discard a functionally relevant term before it can be tested.
\end{rem}

\subsection{Signed support recovery}

Exact unthresholded support recovery requires stronger assumptions than RE. We include the standard condition only to clarify when Lasso alone can recover support.

\begin{thm}[Signed support recovery under irrepresentability]
	\label{thm:signed_support_irrepresentable}
	Let \(\Sigma=A^\top A/n\), and assume \(\Sigma_{SS}\) is nonsingular. Suppose there exists \(\rho\in(0,1]\) such that
	\begin{equation}
		\label{eq:irrepresentable_condition}
		\left\|
		\Sigma_{S^cS}\Sigma_{SS}^{-1}\operatorname{sign}(\beta^\star_S)
		\right\|_\infty
		\le
		1-\rho .
	\end{equation}
	Assume also that the noise satisfies
	\[
	\left\|\frac{1}{n}A^\top\eta\right\|_\infty
	\le
	c_\rho\lambda
	\]
	for a sufficiently small constant \(c_\rho>0\), and that
	\begin{equation}
		\label{eq:beta_min_sign}
		\min_{j\in S}|\beta_j^\star|
		>
		\left\|\Sigma_{SS}^{-1}\right\|_{\infty\to\infty}
		\left(
		\left\|\frac{1}{n}A_S^\top\eta\right\|_\infty+\lambda
		\right).
	\end{equation}
	Then the Lasso solution is sign consistent:
	\[
	\operatorname{sign}(\hat{\beta})
	=
	\operatorname{sign}(\beta^\star),
	\qquad
	\operatorname{supp}(\hat{\beta})=S .
	\]
\end{thm}

\begin{proof}
	This is the standard primal-dual witness theorem for Lasso sign consistency~\cite{zhao2006model}. The irrepresentable condition controls the off-support KKT multipliers, and \eqref{eq:beta_min_sign} prevents sign changes on \(S\).
\end{proof}

\begin{rem}
	Theorem~\ref{thm:signed_support_irrepresentable} is not used as a causal relevance result. It is a finite-design support result for the anchored regression problem. Counterfactual relevance is still decided by the deletion scores \(\delta_j\).
\end{rem}

\subsection{Empirical restricted-eigenvalue condition}

The design matrix in PDE discovery is built from operator evaluations. Its conditioning depends on the trajectory, the sampling distribution, and the library.

\begin{lemma}[Empirical RE under randomized operator sampling]
	\label{lem:empirical_RE}
	Let
	\[
	\psi_j(X,T)=\mathcal{T}_j[u^\dagger](X,T),
	\qquad j=1,\ldots,m,
	\]
	be standardized so that \(\mathbb{E}_\mu[\psi_j^2]=1\). Assume the vector
	\[
	\psi=(\psi_1,\ldots,\psi_m)
	\]
	is uniformly sub-Gaussian with parameter \(K\), and define
	\[
	\Sigma_\mu=\mathbb{E}_\mu[\psi(X,T)\psi(X,T)^\top].
	\]
	Suppose that, for every \(S\subset[m]\) with \(|S|\le2s\),
	\[
	\lambda_{\min}\big((\Sigma_\mu)_{SS}\big)\ge\lambda_0>0 .
	\]
	Draw \((X_i,T_i)\overset{\mathrm{i.i.d.}}{\sim}\mu\), and set \(A_{ij}=\psi_j(X_i,T_i)\). Then there exist constants \(c,C>0\), depending only on \(K\), such that if
	\[
	n\ge
	C\lambda_0^{-2}s\log\!\left(\frac{em}{s\delta}\right),
	\]
	then \(A\) satisfies Assumption~\ref{ass:RE} with
	\[
	\kappa\ge c\lambda_0
	\]
	with probability at least \(1-\delta\).
\end{lemma}

\begin{proof}
	The result is the standard restricted covariance concentration bound for sub-Gaussian designs applied to the operator-feature vector \(\psi(X,T)\). Uniform concentration over \(2s\)-sparse unit vectors follows from a covering argument over sparse supports. The cone extension is the usual reduction from sparse eigenvalues to restricted eigenvalues in Lasso theory~\cite{bickel2009simultaneous}.
\end{proof}

\begin{rem}[Deterministic designs]
	If collocation points are deterministic, the matrix \(A\) is fixed. The relevant diagnostic is then the empirical restricted-eigenvalue constant
	\[
	\widehat{\kappa}(A;s)
	=
	\inf_{\substack{|S|\le s\\ \Delta\neq0,\ \|\Delta_{S^c}\|_1\le3\|\Delta_S\|_1}}
	\frac{\|A\Delta\|_2^2/n}{\|\Delta\|_2^2}.
	\]
	This quantity should be reported for the actual operator-evaluation matrix used in the experiment. The stochastic assumptions in Lemma~\ref{lem:empirical_RE} belong to the sampling design, not to the PDE model.
\end{rem}

\subsection{Reusable screening-to-counterfactual contract}

We now state the reusable guarantee connecting sparse screening with the counterfactual pruning theorem. Let
\[
S_{\cf}^\star=\{j:\delta_j\ge\varepsilon\}
\]
be the exact counterfactual relevance set at tolerance \(\varepsilon\).

\begin{thm}[Screening-to-pruning contract]
	\label{thm:screening_pruning_contract}
	Assume the anchored regression has active support \(S\), and that
	\[
	S_{\cf}^\star\subseteq S .
	\]
	Let \(\widehat{S}_\tau\) be the thresholded Lasso support. Suppose
	\[
	\min_{j\in S}|\beta_j^\star|>\tau+B_s,
	\qquad
	B_s=\frac{4\lambda\sqrt{s}}{\kappa}.
	\]
	Then
	\[
	S_{\cf}^\star\subseteq\widehat{S}_\tau .
	\]
	If, in addition, each estimated counterfactual score satisfies
	\[
	|\widehat{\delta}_j-\delta_j|\le\eta
	\qquad
	\forall j\in\widehat{S}_\tau,
	\]
	and the margin condition
	\[
	|\delta_j-\varepsilon|>\eta
	\qquad
	\forall j\in\widehat{S}_\tau
	\]
	holds, then
	\[
	\widehat{S}_{\cf}
	=
	\{j\in\widehat{S}_\tau:\widehat{\delta}_j\ge\varepsilon\}
	\]
	recovers the exact relevance set:
	\[
	\widehat{S}_{\cf}=S_{\cf}^\star .
	\]
\end{thm}

\begin{proof}
	The beta-min condition and Corollary~\ref{cor:thresholded_support_control} imply \(S\subseteq\widehat{S}_\tau\). Since \(S_{\cf}^\star\subseteq S\), we have \(S_{\cf}^\star\subseteq\widehat{S}_\tau\). The equality \(\widehat{S}_{\cf}=S_{\cf}^\star\) follows from the counterfactual pruning margin argument in Theorem~\ref{thm:cf_pruning_consistency}.
\end{proof}

\begin{rem}[Interpretation]
	Theorem~\ref{thm:screening_pruning_contract} is the practical contract of the framework. Sparse regression must be tuned to avoid missing potentially relevant terms. Counterfactual pruning then removes terms that are selected by residual fitting but have no functional effect at the chosen tolerance.
\end{rem}

\subsection{Design protocol for reusable PDE discovery}

The following protocol is used across synthetic and real-data experiments. Its purpose is to make the operator-evaluation matrix informative enough for sparse screening and to make counterfactual tests interpretable.

\paragraph{D1. Anchor or normalize the PDE relation.}
Use an anchored regression of the form \eqref{eq:anchored_regression}, or an equivalent scale-normalized formulation. Do not apply an unconstrained \(\ell_1\) penalty directly to a homogeneous residual.

\paragraph{D2. Build the operator-evaluation matrix.}
Evaluate every candidate operator on the same factual surrogate or observed field:
\[
A_{ij}=\mathcal{T}_j[u](z_i).
\]
For multiple trajectories, stack all rows into one design matrix.

\paragraph{D3. Normalize columns before screening.}
Set
\[
A_{\cdot j}\leftarrow \frac{A_{\cdot j}}{\|A_{\cdot j}\|_2}.
\]
Normalization removes scale effects between operators, but it does not remove collinearity.

\paragraph{D4. Inspect design conditioning.}
Report at least the coherence
\[
\mu(A)=
\max_{j\neq k}
\frac{|A_{\cdot j}^{\top}A_{\cdot k}|}
{\|A_{\cdot j}\|_2\|A_{\cdot k}\|_2},
\]
and an empirical restricted-eigenvalue proxy such as
\[
\widehat{\kappa}(A;s)
=
\inf_{\substack{|S|\le s\\ \Delta\neq0,\ \|\Delta_{S^c}\|_1\le3\|\Delta_S\|_1}}
\frac{\|A\Delta\|_2^2/n}{\|\Delta\|_2^2}.
\]

\paragraph{D5. Use conservative screening.}
Choose \(\lambda\) and \(\tau\) to favour recall of plausible terms. False positives are acceptable at this stage because they will be tested by counterfactual deletion.

\paragraph{D6. De-bias coefficients on the screened support.}
After screening, refit on the original scale:
\[
\hat{\beta}_{\mathrm{db}}
=
\arg\min_{\operatorname{supp}(\beta)\subseteq\widehat{S}_\tau}
\frac{1}{2n}\|A\beta-y\|_2^2+\gamma\|\beta\|_2^2 .
\]
Use \(\gamma=0\) when the restricted design is well-conditioned, and \(\gamma>0\) when the selected Gram matrix is ill-conditioned.

\paragraph{D7. Apply counterfactual pruning.}
For each \(j\in\widehat{S}_\tau\), delete or perturb the term, solve the intervened system, and compute
\[
\widehat{\delta}_j
=
\|u_\theta-u_{\theta,j}^{\cf}\|.
\]
Report
\[
\widehat{S}_{\cf}
=
\{j\in\widehat{S}_\tau:\widehat{\delta}_j\ge\varepsilon\}.
\]

\paragraph{D8. Report abstentions.}
If the certified margin satisfies
\[
|\widehat{\delta}_j-\varepsilon|\le \eta_f+\eta_{\cf},
\]
do not classify the term as relevant or irrelevant. Report it as unresolved at the chosen tolerance.

\paragraph{D9. Use multi-trajectory excitation when aliasing is detected.}
If \(A\) is poorly conditioned or if operator columns are nearly collinear, add trajectories generated by distinct initial conditions, forcings, boundary conditions, or parameter regimes. The purpose is to excite operator directions that a single trajectory cannot separate.

\paragraph{D10. State the scope of each conclusion.}
Every reported relevance statement should specify the library, experiment class, observation window, norm, and threshold:
\[
(\Gamma,\mathcal{E},\mathcal{O},\|\cdot\|,\varepsilon).
\]
This prevents interpreting a local counterfactual relevance decision as unconditional physical causality.

\section{Operational Use of the Counterfactual Framework}
\label{sec:operational-framework}

The preceding results define counterfactual operator relevance and the conditions under which sparse residual screening can be used safely. We now state the operational consequences used in the experiments. The purpose is not to introduce another discovery objective, but to specify how the theory is applied reproducibly to new PDE systems.

\subsection{Observable-level operator relevance}

State-level relevance may be stronger than what an application requires. In many problems the scientific target is an observable, such as total mass, energy, flux, front speed, or a terminal spatial average. We therefore distinguish full-trajectory relevance from observable relevance.

Let \(\mathcal{O}:V\to\mathbb{R}\) be a Fréchet differentiable observable. For deletion of the \(j\)-th operator, define
\begin{equation}
	\label{eq:observable_cf_deviation}
	\Delta_j^{\mathcal{O}}
	=
	|\mathcal{O}(u)-\mathcal{O}(u_j^{\cf})|.
\end{equation}
The operator \(\mathcal{T}_j\) is observable-relevant at tolerance \(\varepsilon_{\mathcal{O}}>0\) if
\[
\Delta_j^{\mathcal{O}}\ge \varepsilon_{\mathcal{O}}.
\]

\begin{thm}[Adjoint approximation of observable relevance]
	\label{thm:observable_adjoint_relevance}
	Assume the local deletion branch in Assumption~\ref{ass:local_solution_branch}. Let
	\[
	g_j=\alpha_j\mathcal{T}_j[u],
	\qquad
	L_u=\partial_u\mathcal{R}(u;\alpha).
	\]
	Let \(\mathcal{O}:V\to\mathbb{R}\) be twice continuously Fréchet differentiable on a neighbourhood of the deletion branch, and suppose
	\[
	\|D^2\mathcal{O}(w)\|_{\mathcal{L}(V,V;\mathbb{R})}\le H_{\mathcal{O}}
	\]
	on that neighbourhood. Let \(\phi\in Y^*\) solve the adjoint equation
	\begin{equation}
		\label{eq:observable_adjoint}
		L_u^*\phi=D\mathcal{O}(u).
	\end{equation}
	Then, for \(\rho\in[0,1]\),
	\begin{equation}
		\label{eq:observable_first_order}
		\mathcal{O}(u_j(\rho))-\mathcal{O}(u)
		=
		-\rho\langle \phi,g_j\rangle
		+
		R_j^{\mathcal{O}}(\rho),
	\end{equation}
	where
	\begin{equation}
		\label{eq:observable_remainder}
		|R_j^{\mathcal{O}}(\rho)|
		\le
		\frac{\rho^2}{2}\|D\mathcal{O}(u)\|_{V^*}M_j
		+
		\frac{H_{\mathcal{O}}}{2}
		\left(
		\rho\|L_u^{-1}g_j\|_V+\frac{\rho^2M_j}{2}
		\right)^2 .
	\end{equation}
	Here
	\[
	M_j=\sup_{\rho\in[0,1]}\|u_j''(\rho)\|_V.
	\]
\end{thm}

\begin{proof}
	By Theorem~\ref{thm:residual_counterfactual_gap},
	\[
	u_j(\rho)-u
	=
	-\rho L_u^{-1}g_j+r_j(\rho),
	\qquad
	\|r_j(\rho)\|_V\le \frac{\rho^2M_j}{2}.
	\]
	Taylor expansion of \(\mathcal{O}\) at \(u\) gives
	\[
	\mathcal{O}(u_j(\rho))-\mathcal{O}(u)
	=
	D\mathcal{O}(u)[u_j(\rho)-u]
	+
	R_{\mathcal{O}},
	\]
	with
	\[
	|R_{\mathcal{O}}|
	\le
	\frac{H_{\mathcal{O}}}{2}\|u_j(\rho)-u\|_V^2 .
	\]
	Since \(L_u^*\phi=D\mathcal{O}(u)\),
	\[
	D\mathcal{O}(u)[L_u^{-1}g_j]
	=
	\langle \phi,g_j\rangle .
	\]
	Combining these estimates gives \eqref{eq:observable_first_order}--\eqref{eq:observable_remainder}.
\end{proof}

\begin{rem}[Use in experiments]
	Equation~\eqref{eq:observable_first_order} gives a reusable first-order screening score,
	\[
	I_j^{\mathcal{O}}=|\langle \phi,\alpha_j\mathcal{T}_j[u]\rangle|,
	\]
	for observable-level relevance. It is not a substitute for finite counterfactual deletion when the intervention is large. It is a low-cost ranking tool for deciding which terms should be tested first.
\end{rem}

\subsection{Constraint-aware library reduction}

Many PDE systems impose constraints that make some candidate operators vanish or become redundant on the admissible solution class. Discovery should therefore be performed on the effective library induced by the experiment, not on the formal library alone.

Let \(M\subset V\) denote the constraint class containing the observed trajectories. Define the constraint-null sublibrary
\begin{equation}
	\label{eq:constraint_null_library}
	\mathcal{N}_M
	=
	\{j\in[m]: \mathcal{T}_j[u]=0 \ \text{for all } u\in M\}.
\end{equation}

\begin{prop}[Identifiable library modulo constraints]
	\label{prop:library_mod_constraints}
	Assume all observed trajectories lie in \(M\). If \(j\in\mathcal{N}_M\), then the coefficient of \(\mathcal{T}_j\) is not identifiable from residual evaluations on those trajectories. More generally, if
	\[
	\mathcal{T}_j[u]=\mathcal{T}_k[u]
	\qquad \forall u\in M,
	\]
	then only the combined coefficient \(\alpha_j+\alpha_k\) is identifiable on \(M\). Hence operator discovery on constrained data identifies equivalence classes of operators modulo their restrictions to \(M\).
\end{prop}

\begin{proof}
	If \(j\in\mathcal{N}_M\), then \(\alpha_j\mathcal{T}_j[u]=0\) for every observed \(u\in M\), so changing \(\alpha_j\) changes no evaluated residual. If \(\mathcal{T}_j[u]=\mathcal{T}_k[u]\) on \(M\), then
	\[
	\alpha_j\mathcal{T}_j[u]+\alpha_k\mathcal{T}_k[u]
	=
	(\alpha_j+\alpha_k)\mathcal{T}_j[u]
	\]
	for all observed trajectories. Thus \(\alpha_j\) and \(\alpha_k\) are not separately identifiable on \(M\).
\end{proof}

\begin{cor}[Maxwell constraint-null term]
	\label{cor:maxwell_operational}
	For divergence-free Maxwell fields satisfying \(\nabla\cdot E=0\), the operator
	\[
	G(E)=\nabla(\nabla\cdot E)
	\]
	vanishes. Its coefficient is therefore not identifiable from data restricted to the divergence-free Maxwell constraint class~\cite{monk2003finite}.
\end{cor}

\begin{cor}[Projected incompressible flow]
	\label{cor:ns_projected_operational}
	For incompressible flow represented after Leray projection,
	\[
	P\nabla p=0.
	\]
	Thus pressure is not identifiable as an independent sparse term in the projected velocity equation, although it remains essential in the constrained Navier--Stokes formulation~\cite{temam1977navier}.
\end{cor}

\subsection{Scope of common PDE learning objectives}

The framework also clarifies what different learning objectives can and cannot identify.

\begin{prop}[Scope of identification for common objectives]
	\label{prop:scope_common_objectives}
	Fix a candidate library \(\Gamma=\{\mathcal{T}_j\}_{j=1}^m\) and a finite experiment.
	
	\begin{enumerate}
		\item A fixed-operator PINN enforces a prescribed residual \(\mathcal{N}[u]=0\), but it does not estimate the support of \(\Gamma\) unless additional coefficient variables and selection rules are introduced~\cite{raissi_physics-informed_2019}.
		
		\item A DeepONet or neural operator learns a solution map on the training distribution, but it does not determine a unique PDE operator without an additional structural identification step~\cite{lu_learning_2021}.
		
		\item Sparse residual regression estimates a candidate support only after the PDE relation is placed in an identifiable regression form, such as an anchored or scale-normalized design~\cite{rudy_data-driven_2017}.
		
		\item Counterfactual validation tests whether a selected term changes the induced trajectory or observable under a specified intervention. It does not remove the need for sufficient excitation, constraint-aware modelling, and a well-conditioned operator-evaluation design.
	\end{enumerate}
\end{prop}

\begin{proof}
	The first two claims follow from the optimization variables in the respective objectives: neither a fixed-residual PINN loss nor a solution-map loss contains a sparse coefficient estimator over \(\Gamma\). The third claim follows because a homogeneous penalized residual admits the trivial solution unless a scale convention or anchor is imposed. The fourth follows from the definition of the counterfactual score, which is conditional on the fitted model, intervention, experiment, and tolerance.
\end{proof}

\subsection{Reusable discovery algorithm}

The algorithm below implements the screening-to-counterfactual contract. It separates support proposal, coefficient de-biasing, counterfactual deletion, and decision certification.

\begin{algorithm}[H]
	\caption{Counterfactual Operator Discovery}
	\label{alg:counterfactual_operator_discovery}
	\begin{algorithmic}[1]
		\State \textbf{Input:} data \(\mathcal{D}\), library \(\Gamma\), anchor or scale constraint, thresholds \(\tau,\varepsilon\)
		\State Build the anchored operator-evaluation design \(y=A\beta+\eta\)
		\State Normalize columns of \(A\)
		\State Fit a sparse screening model to obtain \(\hat{\beta}\)
		\State Set \(S_{\mathrm{fit}}=\{j:|\hat{\beta}_j|>\tau\}\)
		\State Refit coefficients on \(S_{\mathrm{fit}}\) on the original scale
		\For{\(j\in S_{\mathrm{fit}}\)}
		\State Delete or perturb \(\mathcal{T}_j\)
		\State Solve or retrain the intervened model to obtain \(u_{\theta,j}^{\cf}\)
		\State Compute \(\widehat{\delta}_j=\|u_\theta-u_{\theta,j}^{\cf}\|\)
		\State Optionally compute \(\widehat{\Delta}_j^{\mathcal{O}}=|\mathcal{O}(u_\theta)-\mathcal{O}(u_{\theta,j}^{\cf})|\)
		\State Classify \(j\) using the certified margin in Theorem~\ref{thm:robust_cf_decision}
		\EndFor
		\State \textbf{Return:} screened support \(S_{\mathrm{fit}}\), relevance set \(S_{\cf}\), unresolved set \(S_{\mathrm{abs}}\)
	\end{algorithmic}
\end{algorithm}

The output contains three sets. The screened support \(S_{\mathrm{fit}}\) records terms selected by residual regression. The relevance set \(S_{\cf}\) records terms whose deletion changes the state or observable beyond tolerance. The unresolved set \(S_{\mathrm{abs}}\) records terms whose estimated counterfactual score lies inside the surrogate uncertainty band.

\subsection{Benchmark design as theorem checks}

The validation suite should instantiate the main theoretical claims rather than merely list PDE examples. Table~\ref{tab:benchmark_theorem_checks} gives the reusable benchmark contract.

\begin{table}[!ht]
	\centering
	\caption{Benchmark design aligned with the theoretical results.}
	\label{tab:benchmark_theorem_checks}
	\renewcommand{\arraystretch}{1.2}
	\resizebox{\textwidth}{!}{
		\begin{tabular}{p{3.4cm}p{5.2cm}p{5.8cm}}
			\toprule
			\textbf{Theoretical issue} 
			& \textbf{Benchmark requirement} 
			& \textbf{Reported quantity} \\
			\midrule
			Residual-counterfactual gap 
			& Include a case where residual contribution and deletion response disagree. 
			& \(C_j\), \(\widehat{\delta}_j\), and the linearized score \(\|L_u^{-1}g_j\|\). \\
			\midrule
			Operator aliasing 
			& Include a single-mode or narrow-trajectory case with collinear operator columns. 
			& \(\mu(A)\), rank of \(A_S\), and equivalence of fitted trajectories. \\
			\midrule
			Multi-experiment separability 
			& Add trajectories that excite different operator directions. 
			& Change in \(\widehat{\kappa}(A;s)\), support stability, and deletion scores. \\
			\midrule
			Constraint-null non-identifiability 
			& Include a constrained PDE with an operator that vanishes on the constraint class. 
			& Constraint residual, coefficient instability, and \(\widehat{\delta}_j\). \\
			\midrule
			Certified pruning 
			& Use surrogate residual or validation error to form an uncertainty band. 
			& Relevant, irrelevant, and unresolved sets at tolerance \(\varepsilon\). \\
			\bottomrule
	\end{tabular}}
\end{table}

This benchmark design prevents overclaiming. Exact support recovery should be reported only for synthetic systems with known active support. Real-data studies should report predictive error, design diagnostics, support stability, counterfactual deviations, and unresolved terms.

\subsection{Reporting contract}

Each experiment should report the tuple
\[
(\Gamma,\mathcal{E},\mathcal{O},\|\cdot\|,\varepsilon),
\]
where \(\Gamma\) is the library, \(\mathcal{E}\) is the experiment class, \(\mathcal{O}\) is the observation or target functional, \(\|\cdot\|\) is the decision norm, and \(\varepsilon\) is the relevance threshold. The reported operator set is therefore not an unconditional physical law. It is a counterfactual relevance set for the stated experimental and modelling conditions.

\begin{table}[!ht]
	\centering
	\caption{Minimum reporting items for reusable counterfactual PDE discovery.}
	\label{tab:reporting_contract}
	\renewcommand{\arraystretch}{1.2}
		\begin{tabular}{p{4.1cm}p{8.8cm}}
			\toprule
			\textbf{Item} & \textbf{Required report} \\
			\midrule
			Library & All candidate operators, including removed or grouped terms. \\
			Experiment class & Initial conditions, boundary conditions, forcings, parameter regimes, and sampling scheme. \\
			Design diagnostics & Column coherence, empirical restricted-eigenvalue proxy, selected Gram condition number. \\
			Screening rule & Anchor, normalization, \(\lambda\), threshold \(\tau\), and de-biasing method. \\
			Counterfactual rule & Deleted or perturbed term, refitting policy, decision norm, threshold \(\varepsilon\). \\
			Uncertainty & Surrogate residual, validation error, certified margin, or bootstrap interval. \\
			Output & \(S_{\mathrm{fit}}\), \(S_{\cf}\), unresolved terms, and counterfactual scores. \\
			\bottomrule
	\end{tabular}
\end{table}

\section{Experimental Validation}
\label{sec:experiments}

The experiments are designed as checks of the preceding theory. We do not use the validation section to claim unconditional recovery of physical laws. In synthetic experiments, the active support is known and support recovery can be evaluated directly. In real-data case studies, where the true PDE operator is unavailable, we report predictive error, design conditioning, support stability, counterfactual deviations, and unresolved terms.

The validation has two regimes. Controlled synthetic experiments test theorem-level mechanisms such as residual--counterfactual gaps, aliasing, multi-trajectory separability, and constraint-null non-identifiability. Public geophysical fields then test whether the procedure gives stable and interpretable operator surrogates on real spatiotemporal data. The real-data runs use a public atmospheric reanalysis temperature field over the Uganda/East Africa window, denoted \texttt{era5} in the output files, and NOAA OISST sea-surface temperature over the western Indian Ocean/East African coast~\cite{kalnay1996ncep,reynolds2007daily}.

The real-data runs use a public atmospheric reanalysis temperature field over the Uganda/East Africa window, denoted \texttt{era5} in the output files, and NOAA OISST sea-surface temperature over the western Indian Ocean/East African coast~\cite{kalnay1996ncep,reynolds2007daily}. For each experiment we report the operator library, design diagnostics, fitted support, residual contribution index, and one-step counterfactual deletion score.

\subsection{Main validation outputs}

\begin{figure*}[!t]
	\centering
	\begin{subfigure}{0.32\textwidth}
		\centering
		\includegraphics[width=\linewidth]{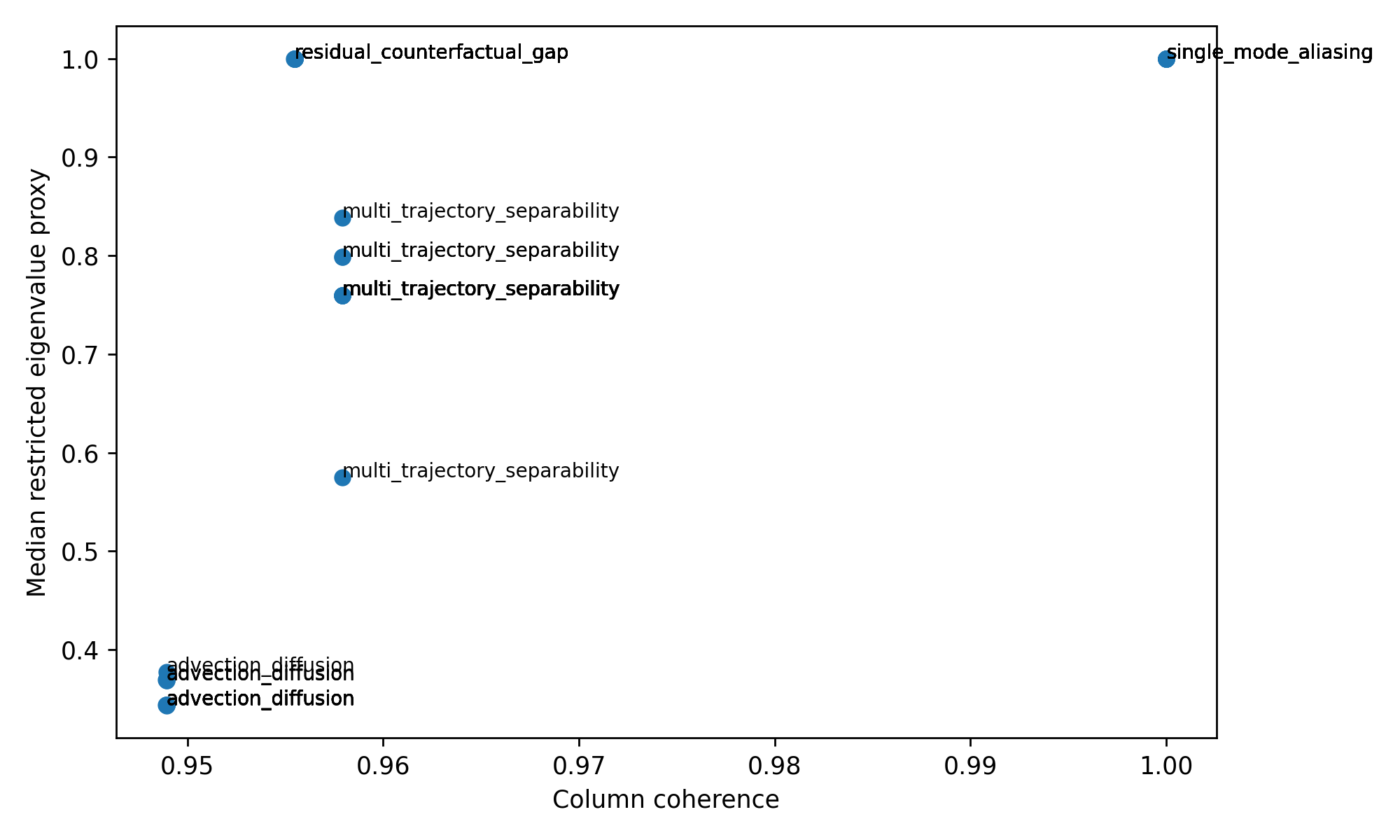}
		\caption{Synthetic design diagnostics.}
		\label{fig:design_synthetic}
	\end{subfigure}
	\hfill
	\begin{subfigure}{0.32\textwidth}
		\centering
		\includegraphics[width=\linewidth]{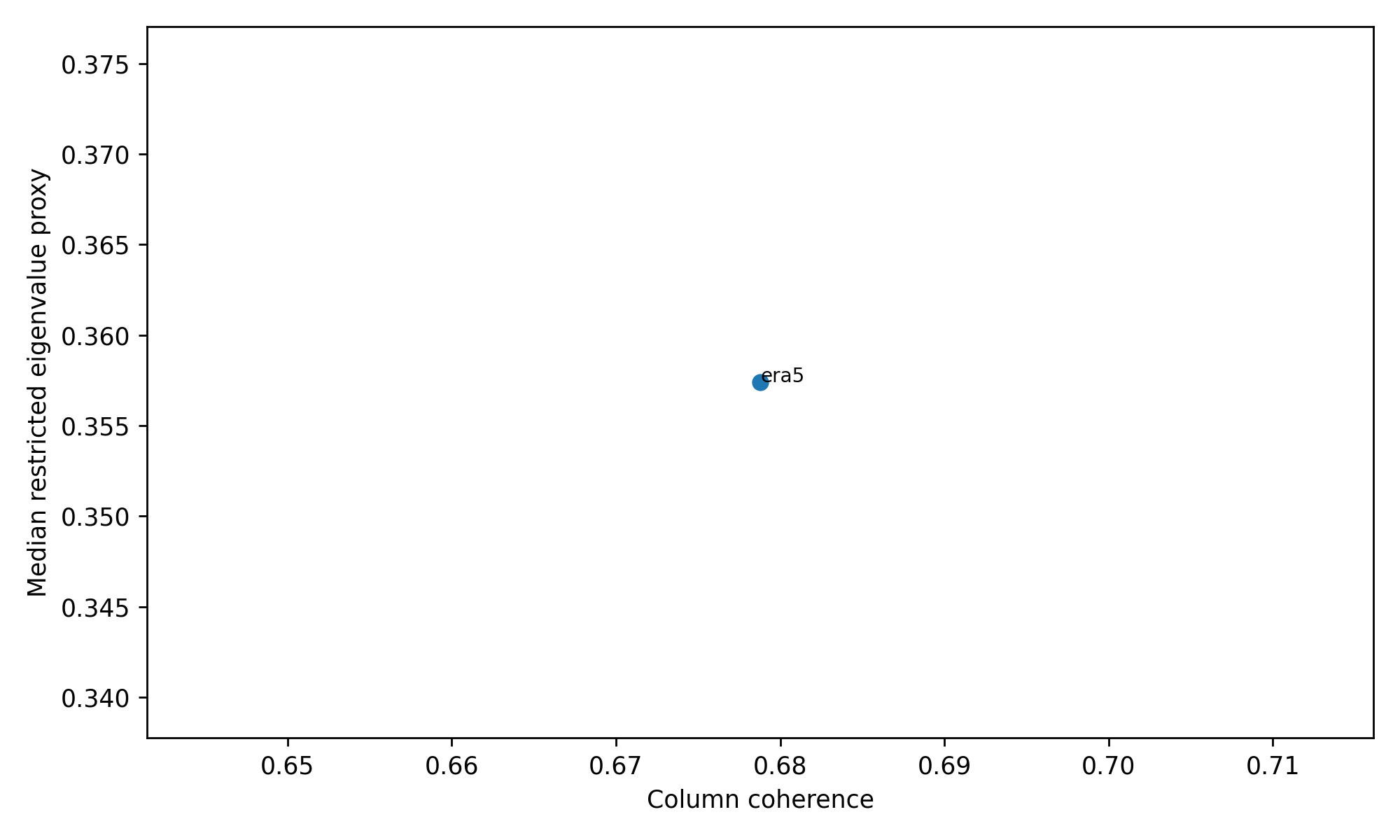}
		\caption{Atmospheric reanalysis design diagnostics.}
		\label{fig:design_era5}
	\end{subfigure}
	\hfill
	\begin{subfigure}{0.32\textwidth}
		\centering
		\includegraphics[width=\linewidth]{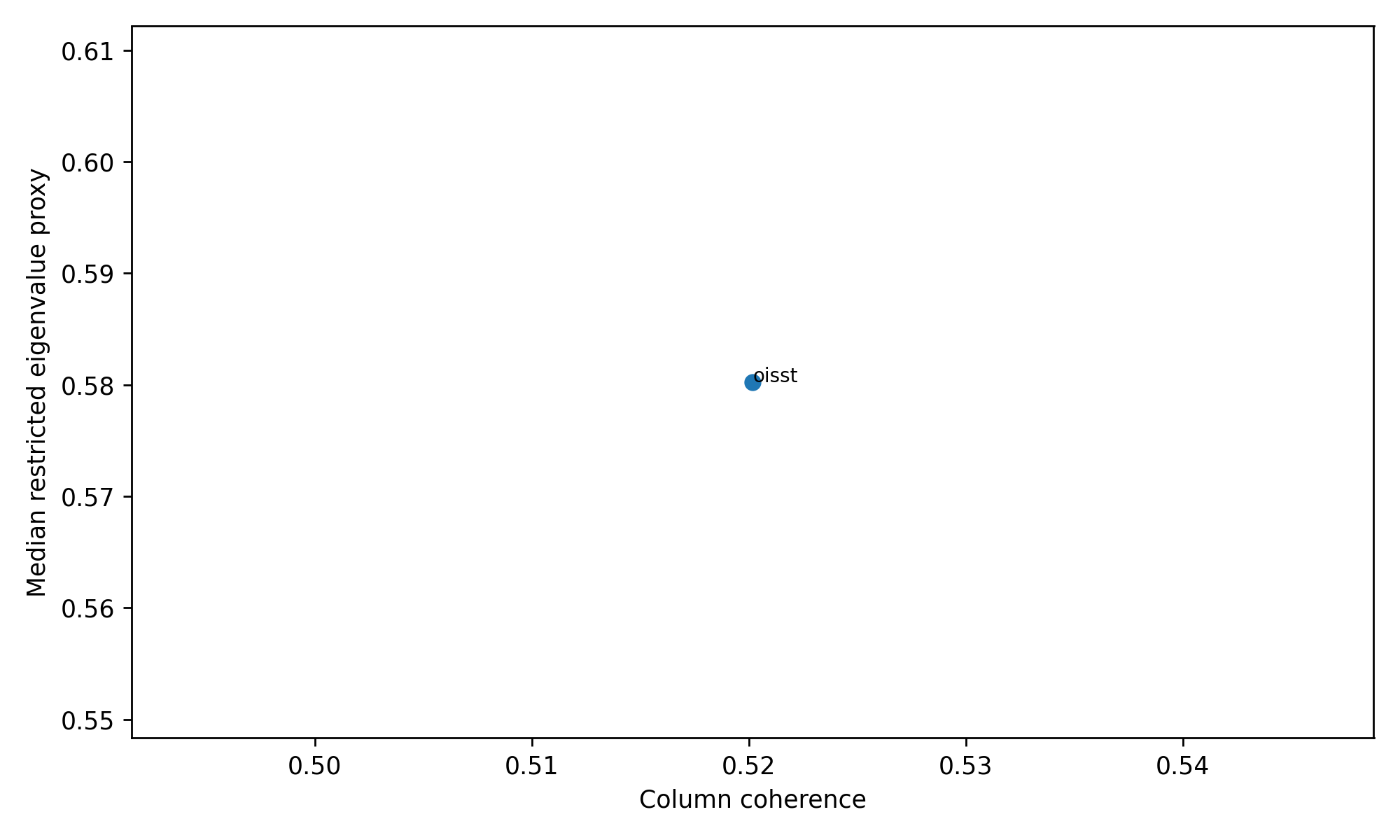}
		\caption{OISST design diagnostics.}
		\label{fig:design_oisst}
	\end{subfigure}
	
	\vspace{0.75em}
	
	\begin{subfigure}{0.32\textwidth}
		\centering
		\includegraphics[width=\linewidth]{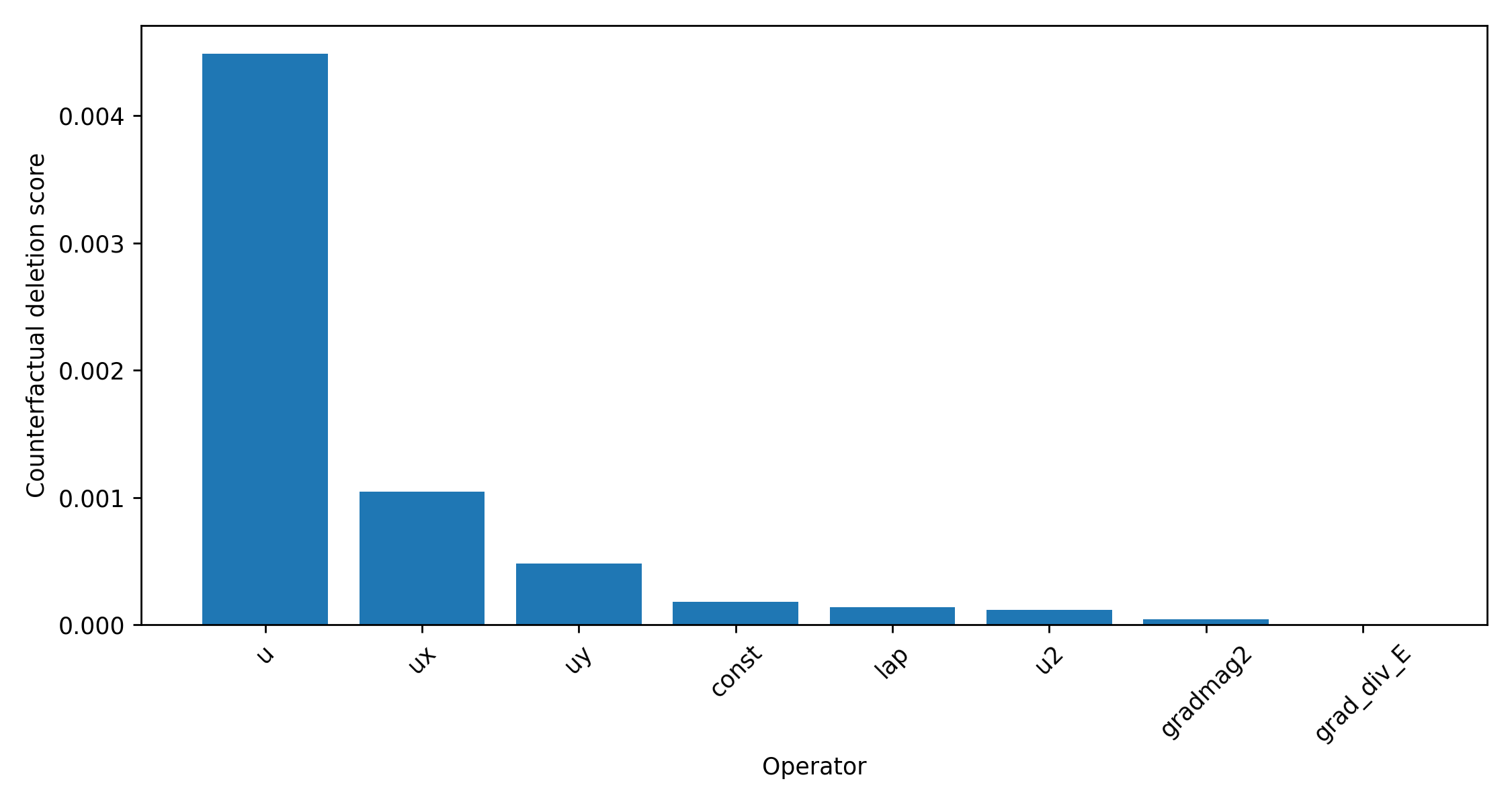}
		\caption{Synthetic operator relevance.}
		\label{fig:relevance_synthetic}
	\end{subfigure}
	\hfill
	\begin{subfigure}{0.32\textwidth}
		\centering
		\includegraphics[width=\linewidth]{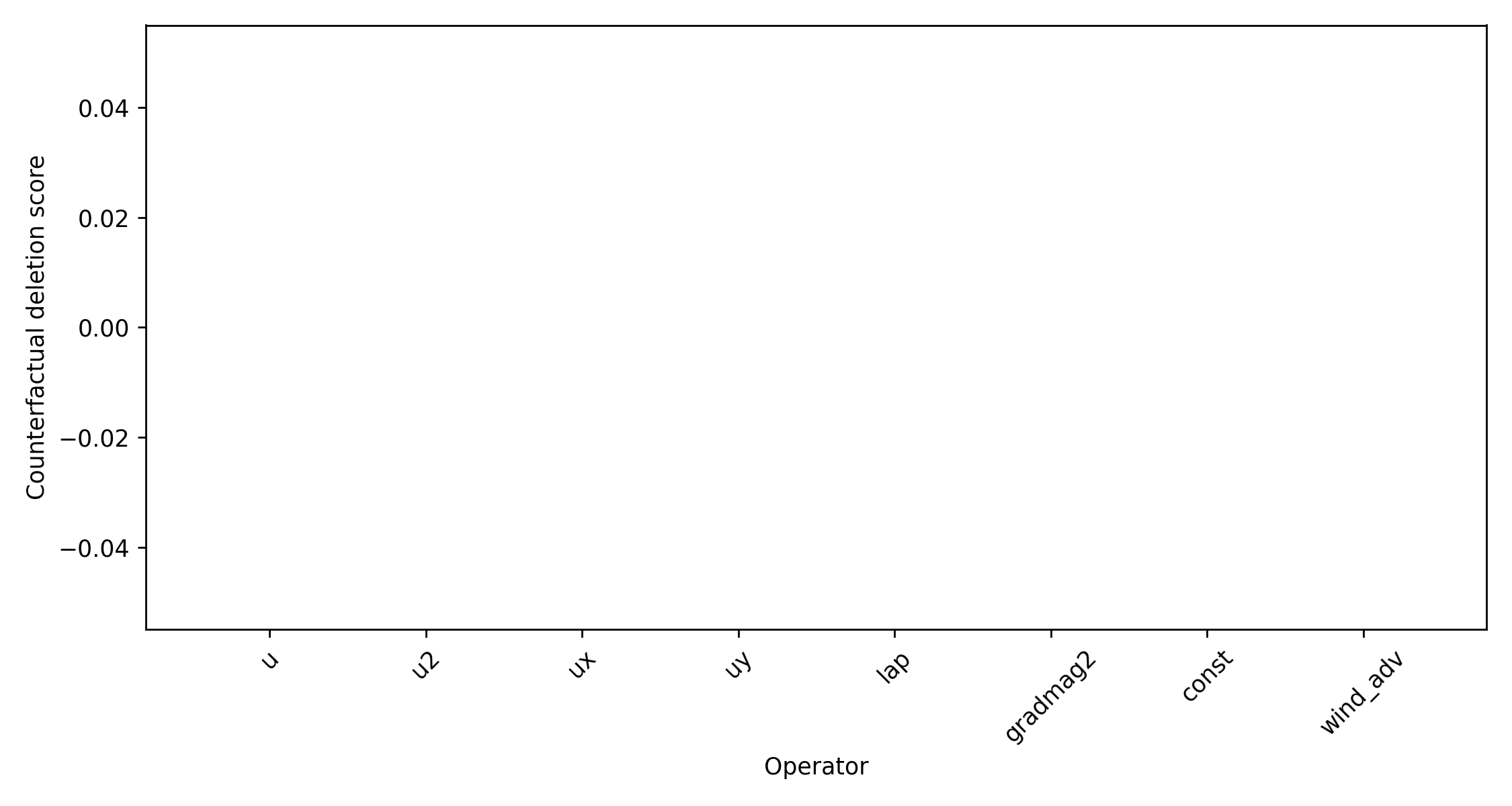}
		\caption{Atmospheric reanalysis operator relevance.}
		\label{fig:relevance_era5}
	\end{subfigure}
	\hfill
	\begin{subfigure}{0.32\textwidth}
		\centering
		\includegraphics[width=\linewidth]{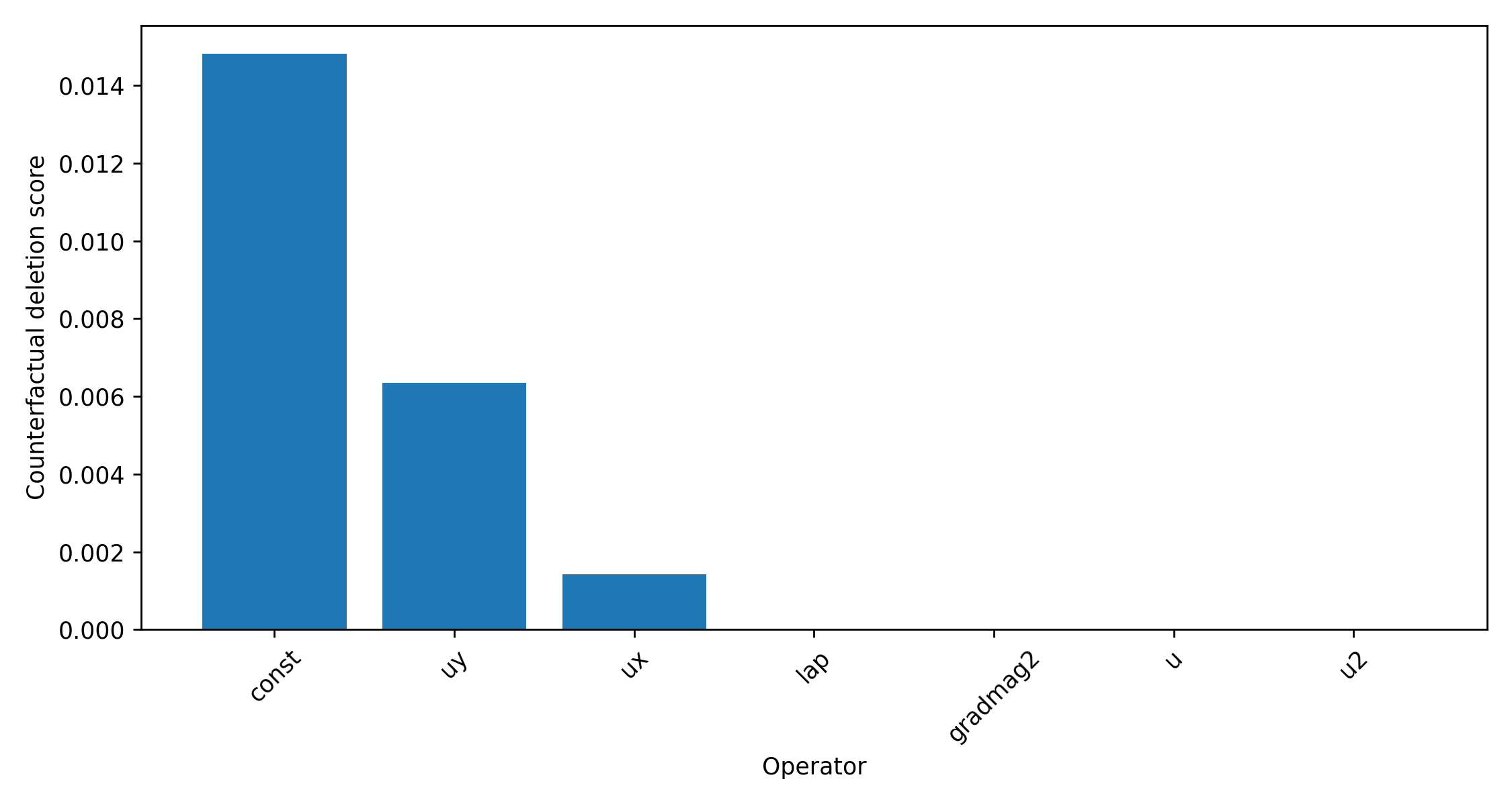}
		\caption{OISST operator relevance.}
		\label{fig:relevance_oisst}
	\end{subfigure}
	\caption{Main validation diagnostics. The top row reports conditioning of the operator-evaluation designs. The bottom row reports one-step counterfactual deletion scores. The synthetic and OISST cases produce interpretable deletion structure, while the atmospheric reanalysis run is inconclusive because no operator was selected by the sparse screen.}
	\label{fig:main_validation_panels}
\end{figure*}

Figure~\ref{fig:main_validation_panels} summarizes the main visual outputs. Panels~\ref{fig:main_validation_panels}a--\ref{fig:main_validation_panels}c show the design diagnostics, while Panels~\ref{fig:main_validation_panels}d--\ref{fig:main_validation_panels}f show the corresponding operator-deletion relevance scores. The synthetic panels are used to check the theoretical mechanisms. The atmospheric reanalysis panel is deliberately interpreted as inconclusive because the sparse screen selected no active operator. The OISST panel provides the clearest real-data operator-surrogate result.

Table~\ref{tab:validation_summary} gives the run-level outputs. The synthetic experiments are summarized by mean selected rate, mean deletion score, and mean residual contribution index across operators. The real-data studies are summarized by the derivative-prediction error, baseline error, skill against the mean predictor, grid spacing, and time step. The atmospheric reanalysis run used \(174\) samples and eight candidate operators, but its skill against the mean predictor was slightly negative. The OISST run used \(250{,}000\) samples and seven candidate operators, with a small positive skill of \(0.002459\).

\begin{table*}[!t]
	\centering
	\caption{Run-level validation summary. Synthetic rows report operator-level averages across the benchmark task. Real-data rows report derivative-prediction error and grid metadata.}
	\label{tab:validation_summary}
	\renewcommand{\arraystretch}{1.18}
	\resizebox{\textwidth}{!}{
		\begin{tabular}{lcccccccc}
			\toprule
			\textbf{Case}
			& \textbf{Samples}
			& \textbf{Operators}
			& \textbf{RMSE}
			& \textbf{Baseline RMSE}
			& \textbf{Skill}
			& \textbf{Mean deletion}
			& \textbf{Mean RCI}
			& \textbf{Grid / time step}
			\\
			\midrule
			Atmospheric reanalysis
			& \(174\)
			& \(8\)
			& \(6.049\times 10^{-6}\)
			& \(6.015\times 10^{-6}\)
			& \(-5.590\times 10^{-3}\)
			& --
			& --
			& \(208.46\text{ km}\times211.80\text{ km};\ 86400\text{ s}\)
			\\
			OISST
			& \(250000\)
			& \(7\)
			& \(1.096\times 10^{-6}\)
			& \(1.099\times 10^{-6}\)
			& \(2.459\times 10^{-3}\)
			& --
			& --
			& \(27.80\text{ km}\times27.80\text{ km};\ 86400\text{ s}\)
			\\
			Residual--counterfactual gap
			& \(151008\)
			& \(7\)
			& --
			& --
			& --
			& \(9.057\times 10^{-4}\)
			& \(1.438\times 10^{-1}\)
			& synthetic
			\\
			Single-mode aliasing
			& \(151008\)
			& \(7\)
			& --
			& --
			& --
			& \(8.049\times 10^{-4}\)
			& \(1.429\times 10^{-1}\)
			& synthetic
			\\
			Multi-trajectory separability
			& \(151008\)
			& \(7\)
			& --
			& --
			& --
			& \(9.024\times 10^{-4}\)
			& \(1.699\times 10^{-1}\)
			& synthetic
			\\
			Advection--diffusion
			& \(151008\)
			& \(7\)
			& --
			& --
			& --
			& \(1.095\times 10^{-3}\)
			& \(2.427\times 10^{-1}\)
			& synthetic
			\\
			Constraint-null
			& \(64000\)
			& \(1\)
			& --
			& --
			& --
			& \(3.012\times 10^{-15}\)
			& \(1.329\times 10^{-13}\)
			& synthetic
			\\
			\bottomrule
	\end{tabular}}
\end{table*}

\subsection{Design diagnostics}

Table~\ref{tab:design_diagnostics_main} reports the empirical conditioning of the operator-evaluation matrices. The OISST design is better conditioned than the atmospheric reanalysis design, with lower coherence \(0.5202\), larger median restricted eigenvalue proxy \(0.5802\), and smaller median selected-support condition number \(2.4457\). The atmospheric reanalysis design has only \(174\) rows and a coarser grid, with coherence \(0.6788\) and median restricted eigenvalue proxy \(0.3574\). This conditioning, together with the zero selected support in Table~\ref{tab:real_operator_relevance_oisst}, explains why the atmospheric case is treated as unresolved rather than as negative evidence against the method.

\begin{table*}[!t]
	\centering
	\caption{Design diagnostics for synthetic and real-data operator-evaluation matrices. Synthetic values are means over ten seeds where applicable.}
	\label{tab:design_diagnostics_main}
	\renewcommand{\arraystretch}{1.18}
	\resizebox{\textwidth}{!}{
		\begin{tabular}{lccccc}
			\toprule
			\textbf{Case}
			& \textbf{Rows}
			& \textbf{Operators}
			& \textbf{Coherence}
			& \textbf{Median restricted eigenvalue}
			& \textbf{Median condition number}
			\\
			\midrule
			Atmospheric reanalysis
			& \(174\)
			& \(8\)
			& \(0.6788\)
			& \(0.3574\)
			& \(4.9882\)
			\\
			OISST
			& \(250000\)
			& \(7\)
			& \(0.5202\)
			& \(0.5802\)
			& \(2.4457\)
			\\
			Residual--counterfactual gap
			& \(151008\)
			& \(7\)
			& \(0.9555\)
			& \(1.0000\)
			& \(1.0000\)
			\\
			Single-mode aliasing
			& \(151008\)
			& \(7\)
			& \(1.0000\)
			& \(1.0000\)
			& \(1.0000\)
			\\
			Multi-trajectory separability
			& \(151008\)
			& \(7\)
			& \(0.9579\)
			& \(0.7569\)
			& \(1.6682\)
			\\
			Advection--diffusion
			& \(151008\)
			& \(7\)
			& \(0.9489\)
			& \(0.3546\)
			& \(4.6491\)
			\\
			Constraint-null
			& \(64000\)
			& \(1\)
			& \(0.0000\)
			& --
			& --
			\\
			\bottomrule
	\end{tabular}}
\end{table*}

The synthetic diagnostics show the intended theory checks. The single-mode aliasing experiment has perfect coherence \(1.0000\), matching the aliasing theorem. The multi-trajectory separability experiment reduces the practical degeneracy of the design, with a median restricted eigenvalue proxy of \(0.7569\), while the advection--diffusion design is less well conditioned, with median restricted eigenvalue proxy \(0.3546\) and condition number \(4.6491\). The constraint-null case has one candidate operator and therefore no nontrivial restricted spectrum.

\subsection{Synthetic theorem checks}

Table~\ref{tab:synthetic_operator_checks} reports the main operator-level synthetic checks. These values should be read as diagnostics, not as blanket support-recovery claims. In the residual--counterfactual gap experiment, the true \(u\) term is selected in all seeds and has deletion score \(6.310\times10^{-3}\), while the spurious Laplacian term has selection rate \(0\) and deletion score \(3.000\times10^{-5}\). In the single-mode aliasing experiment, the active \(u\) term is selected in all seeds, while the Laplacian term has zero coefficient and zero deletion score under the fitted representation. In the multi-trajectory experiment, \(u\) remains dominant with deletion score \(5.075\times10^{-3}\), while the true but weak Laplacian term has a smaller deletion score \(4.326\times10^{-4}\) and is not selected by the present threshold. This is an important limitation of the current screening setting, and it motivates the recall-oriented screening rule used in the theory.

\begin{table*}[!t]
	\centering
	\caption{Selected synthetic operator-level checks. Values are means over ten seeds. The table reports selected terms and theory-relevant distractors rather than the full operator library.}
	\label{tab:synthetic_operator_checks}
	\renewcommand{\arraystretch}{1.18}
	\resizebox{\textwidth}{!}{
		\begin{tabular}{llccccc}
			\toprule
			\textbf{Experiment}
			& \textbf{Operator}
			& \textbf{True active}
			& \textbf{Selection rate}
			& \textbf{Mean coefficient}
			& \textbf{Mean deletion score}
			& \textbf{Mean RCI}
			\\
			\midrule
			Residual--counterfactual gap
			& \(u\)
			& yes
			& \(1.0000\)
			& \(0.601152\)
			& \(6.310\times10^{-3}\)
			& \(1.001989\)
			\\
			Residual--counterfactual gap
			& \(\Delta u\)
			& no
			& \(0.0000\)
			& \(3.800\times10^{-5}\)
			& \(3.000\times10^{-5}\)
			& \(4.764\times10^{-3}\)
			\\
			Single-mode aliasing
			& \(u\)
			& yes
			& \(1.0000\)
			& \(0.600006\)
			& \(5.634\times10^{-3}\)
			& \(1.000000\)
			\\
			Single-mode aliasing
			& \(\Delta u\)
			& no
			& \(0.0000\)
			& \(0.000000\)
			& \(0.000000\)
			& \(0.000000\)
			\\
			Multi-trajectory separability
			& \(u\)
			& yes
			& \(1.0000\)
			& \(0.489170\)
			& \(5.075\times10^{-3}\)
			& \(0.955393\)
			\\
			Multi-trajectory separability
			& \(\Delta u\)
			& yes
			& \(0.0000\)
			& \(8.820\times10^{-4}\)
			& \(4.326\times10^{-4}\)
			& \(8.144\times10^{-2}\)
			\\
			Advection--diffusion
			& \(u_x\)
			& yes
			& \(1.0000\)
			& \(-0.158608\)
			& \(4.038\times10^{-3}\)
			& \(0.894779\)
			\\
			Advection--diffusion
			& \(u_y\)
			& yes
			& \(1.0000\)
			& \(0.080316\)
			& \(1.914\times10^{-3}\)
			& \(0.424240\)
			\\
			Advection--diffusion
			& \(\Delta u\)
			& yes
			& \(0.0000\)
			& \(2.940\times10^{-4}\)
			& \(9.574\times10^{-5}\)
			& \(2.122\times10^{-2}\)
			\\
			Constraint-null
			& \(\nabla(\nabla\cdot E)\)
			& no
			& \(0.0000\)
			& \(1.000000\)
			& \(3.012\times10^{-15}\)
			& \(1.329\times10^{-13}\)
			\\
			\bottomrule
	\end{tabular}}
\end{table*}

The constraint-null experiment is the cleanest check of Theorem~\ref{thm:constraint_null_nonidentifiability}. The coefficient assigned to \(\nabla(\nabla\cdot E)\) is immaterial because the deletion score is \(3.012\times10^{-15}\) and the residual contribution index is \(1.329\times10^{-13}\). This matches the theory: an operator that vanishes on the admissible constraint class cannot be identified from trajectories restricted to that class.

\subsection{Real-data operator relevance}

The real-data studies are deliberately interpreted as operator-surrogate diagnostics. The atmospheric reanalysis run selected no active operators: all eight candidate terms had coefficient \(0\), residual contribution \(0\), deletion score \(0\), and decision \texttt{unresolved}. This is reported as an inconclusive coarse-grid case, not as evidence of physical absence.

The OISST case gives a more informative real-data result. Table~\ref{tab:real_operator_relevance_oisst} reports all seven candidate operators. The sparse screen selected \(u_x\), \(u_y\), \(\Delta u\), \(|\nabla u|^2\), and the constant term. Counterfactual deletion retained \(u_y\) and the constant term as relevant at the reported threshold \(\varepsilon=0.002412\) with margin \(0.000121\). The meridional gradient term \(u_y\) has deletion score \(6.342\times10^{-3}\), while the constant term has deletion score \(1.481\times10^{-2}\). By contrast, \(u_x\), \(\Delta u\), and \(|\nabla u|^2\) were selected by sparse fitting but had deletion scores below threshold, illustrating the screening-to-pruning distinction.

\begin{table*}[!t]
	\centering
	\caption{OISST operator relevance table. The relevance threshold is \(\varepsilon=0.002412\), with margin \(0.000121\).}
	\label{tab:real_operator_relevance_oisst}
	\renewcommand{\arraystretch}{1.18}
	\resizebox{\textwidth}{!}{
		\begin{tabular}{lcccccc}
			\toprule
			\textbf{Operator}
			& \textbf{Coefficient}
			& \textbf{Selected}
			& \textbf{Residual contribution}
			& \textbf{RCI}
			& \textbf{Deletion score}
			& \textbf{Decision}
			\\
			\midrule
			\(u\)
			& \(0.000000\)
			& no
			& \(0.000000\)
			& \(0.000000\)
			& \(0.000000\)
			& irrelevant
			\\
			\(u^2\)
			& \(0.000000\)
			& no
			& \(0.000000\)
			& \(0.000000\)
			& \(0.000000\)
			& irrelevant
			\\
			\(u_x\)
			& \(8.624\times10^{-3}\)
			& yes
			& \(8.274\times10^{-6}\)
			& \(8.826\times10^{-2}\)
			& \(1.430\times10^{-3}\)
			& irrelevant
			\\
			\(u_y\)
			& \(4.765\times10^{-2}\)
			& yes
			& \(3.670\times10^{-5}\)
			& \(3.915\times10^{-1}\)
			& \(6.342\times10^{-3}\)
			& relevant
			\\
			\(\Delta u\)
			& \(-1.310\times10^{-5}\)
			& yes
			& \(4.441\times10^{-13}\)
			& \(4.737\times10^{-9}\)
			& \(7.674\times10^{-11}\)
			& irrelevant
			\\
			\(|\nabla u|^2\)
			& \(-8.872\times10^{-6}\)
			& yes
			& \(5.164\times10^{-14}\)
			& \(5.509\times10^{-10}\)
			& \(8.924\times10^{-12}\)
			& irrelevant
			\\
			constant
			& \(-1.714\times10^{-7}\)
			& yes
			& \(8.570\times10^{-5}\)
			& \(9.142\times10^{-1}\)
			& \(1.481\times10^{-2}\)
			& relevant
			\\
			\bottomrule
	\end{tabular}}
\end{table*}

The OISST result demonstrates the practical role of counterfactual pruning. Sparse screening alone would retain five terms. Counterfactual deletion reduces this to two terms at the stated tolerance. This is exactly the distinction formalized by Theorem~\ref{thm:screening_pruning_contract}: sparse residual fitting proposes a candidate support, while deletion scores determine functional relevance for the specified experiment and norm.

\subsection{Summary of validation evidence}

The validation supports four claims. First, the synthetic experiments reproduce the intended failure modes: aliasing, weak operator separability, and constraint-null non-identifiability. Second, the design diagnostics show that conditioning is experiment-dependent and must be reported before interpreting sparse support. Third, the atmospheric reanalysis case shows that the framework can abstain when the data and design do not support a meaningful operator decision. Fourth, the OISST case shows the intended screening-to-pruning behaviour on real data: several operators are selected by sparse fitting, but only \(u_y\) and the constant term exceed the counterfactual relevance threshold. These results are sufficient for main-text validation; detailed bootstrap stability paths, full operator tables, and additional diagnostics should be placed in the appendix.

\subsection{Discussion}
\label{subsec:discussion}

The results support the main purpose of the paper, which is to separate residual support from counterfactual operator relevance. The theory shows that an operator can contribute to a residual while having limited effect on the induced trajectory, because the relevant object is the propagated response through the inverse linearized operator, not the residual magnitude alone. This distinction is visible in the synthetic experiments and in the OISST real-data case. Sparse screening proposes candidate operators, while counterfactual deletion determines which selected terms have measurable functional effect under the stated norm and tolerance.

The synthetic experiments provide controlled checks of the theoretical mechanisms. The single-mode experiment reproduces the aliasing obstruction in Theorem~\ref{thm:aliasing_obstruction}, since the design becomes perfectly collinear, as reflected by the coherence value \(1.0000\) in Table~\ref{tab:design_diagnostics_main}. The constraint-null experiment directly supports Theorem~\ref{thm:constraint_null_nonidentifiability}. The operator \(\nabla(\nabla\cdot E)\) has deletion score \(3.012\times10^{-15}\) and residual contribution index \(1.329\times10^{-13}\), which is numerically consistent with an operator that vanishes on the constraint class. These cases are useful because they show that some discovery failures are not optimization failures. They are identifiability failures induced by the experiment, the constraint class, or the operator library.

The multi-trajectory and advection--diffusion experiments show both the value and the limits of sparse screening. Dominant active terms are recovered reliably, but weak terms such as \(\Delta u\) can remain below the present selection threshold even when they are part of the data-generating equation, as shown in Table~\ref{tab:synthetic_operator_checks}. This is not a contradiction of the framework. It illustrates why the paper treats sparse regression as a screening mechanism rather than as a final relevance test. The practical implication is that screening should be tuned for recall when the goal is counterfactual pruning. A conservative sparse screen may miss weak but functionally relevant operators before the deletion test is applied.

The real-data experiments serve a different purpose from the synthetic checks. They do not provide ground-truth operator recovery, since the true physical PDE governing atmospheric temperature or sea-surface temperature is not available in the candidate library. Instead, they test whether the proposed pipeline produces stable, interpretable, low-dimensional operator surrogates from public scientific fields. The atmospheric reanalysis experiment is appropriately unresolved. It used only \(174\) samples on a coarse grid, had weaker design conditioning, selected no stable operator, and achieved slightly negative skill against the mean predictor in Table~\ref{tab:validation_summary}. This outcome is important because the framework is allowed to abstain when the design does not support a reliable operator decision.

The OISST experiment gives the clearest real-data evidence. The design matrix is better conditioned than the atmospheric case, with lower coherence and a larger restricted-eigenvalue proxy in Table~\ref{tab:design_diagnostics_main}. Sparse screening retained five terms, but counterfactual deletion retained only \(u_y\) and the constant term at the stated threshold, as shown in Table~\ref{tab:real_operator_relevance_oisst}. This supports the screening-to-pruning logic in Theorem~\ref{thm:screening_pruning_contract}. Coefficient stability alone is not the same as counterfactual relevance. The appendix makes this clear because \(u_x\) has a nonzero bootstrap interval in Table~\ref{tab:appendix_bootstrap_intervals}, yet its deletion score remains below the relevance threshold in the main-text counterfactual analysis.

The constant term in the OISST surrogate should be interpreted cautiously. It should not be read as a physical constant forcing law. In this design it represents a low-dimensional background tendency left after the chosen anomaly transformation, spatial derivatives, and finite observation window. Its relevance score indicates that removing it changes the fitted short-horizon surrogate. It does not imply that the ocean dynamics are governed by a constant source term. This distinction is consistent with the reporting contract in Table~\ref{tab:reporting_contract}, where every relevance claim is conditional on the library, experiment class, observation window, norm, and threshold.

Taken together, the validation supports a narrower but stronger claim than full PDE discovery from data. The paper provides a reusable theory and workflow for deciding whether selected candidate operators are functionally relevant within a specified experimental design. The results also show why design diagnostics, aliasing checks, constraint-aware libraries, and abstention regions should be reported alongside any discovered operator set.

\subsubsection{Limitations}
\label{subsubsec:limitations}

The framework depends on the candidate library. If a physically important mechanism is absent from the library, the method can only identify the best available surrogate within the supplied terms. The real-data experiments should therefore be interpreted as low-dimensional operator-surrogate studies, not as complete recovery of atmospheric or oceanic governing laws.

The counterfactual deletion score depends on the chosen intervention. Removing a term, perturbing a coefficient, changing a forcing, and altering a boundary condition are different interventions. They may produce different relevance rankings. For this reason, every reported relevance set must specify the intervention class and the norm used to measure deviation.

The sparse screening stage remains sensitive to finite-sample conditioning. High coherence, weak excitation, and narrow trajectories can hide weak active terms or make different terms observationally indistinguishable. The synthetic experiments deliberately expose this issue. In practical use, multiple initial conditions, forcings, regimes, or observation windows may be needed before the operator-evaluation design is informative enough for screening.

The present implementation uses finite-difference derivative estimates and one-step deletion scores for real-data experiments. These choices make the validation reproducible and transparent, but they are not the only possible implementation. More accurate derivative estimation, weak-form regression, adjoint solves, or full counterfactual retraining could improve numerical fidelity, especially for noisy or irregularly sampled data.

The real-data validation uses publicly available gridded fields, but these fields are already products of data assimilation, interpolation, and model-based processing. This does not invalidate their use as scientific spatiotemporal fields, but it limits the strength of any claim about physical law discovery. The appropriate claim is that the method produces and diagnoses operator surrogates on observationally grounded fields.

The abstention mechanism is essential. When the estimated counterfactual score lies near the decision threshold, or when the design is poorly conditioned, the method should report an unresolved term rather than force a relevance decision. This is a limitation in the sense that the method may decline to decide, but it is also a safeguard against overinterpretation.

\section{Conclusion}
\label{sec:conclusion}

This paper developed a counterfactual theory of operator relevance for PDE discovery. The central distinction is between residual usefulness and functional necessity. Sparse residual fitting can propose candidate terms, but it does not by itself show that a term materially affects the induced trajectory or observable. Counterfactual deletion provides the missing diagnostic by testing the effect of removing or perturbing a selected operator.

The main theoretical results formalize this distinction. The residual-counterfactual gap theorem shows that operator relevance is governed by the inverse linearized PDE response, not by residual magnitude alone. The robust decision theorem gives explicit relevance, irrelevance, and abstention regions under surrogate error. The aliasing and constraint-null results identify settings where operator recovery is impossible from the available experiment. The screening-to-pruning theorem connects sparse residual regression with counterfactual relevance under a recall and margin condition. The observable-level result extends the framework beyond full-state deviations to scientific quantities of interest.

The validation supports these claims at two levels. Synthetic experiments verify the key theory mechanisms under known support, including aliasing, constraint-null non-identifiability, and weak-term screening limitations. Public real-data experiments show how the framework behaves on atmospheric and oceanic scientific fields. The atmospheric case is unresolved, which is the appropriate outcome for a weak finite design. The OISST case demonstrates the intended behaviour of the workflow, where sparse screening proposes several terms and counterfactual deletion retains only those with measurable functional effect at the chosen threshold.

The resulting contribution is not another claim of automatic PDE discovery from data. It is a more cautious and reusable framework for reporting what a discovered operator set actually means. A relevance statement is valid only relative to a library, experiment class, observation window, norm, threshold, and intervention. This makes the framework useful both as a diagnostic layer for existing PDE-discovery methods and as a reporting standard for future work on interpretable scientific machine learning.

\bibliographystyle{ieeetr}
\bibliography{refs}

\appendix

\section{Worked Examples: From Operator Screening to Counterfactual Pruning}
\label{app:worked-examples}

This appendix gives explicit examples showing how the proposed theory is used in practice. The examples are deliberately simple. Their purpose is not to add new numerical evidence, but to show how the same workflow recovers coefficients, detects aliasing, and separates sparse screening from counterfactual relevance.

\subsection{Anchored coefficient recovery in a reaction--diffusion model}
\label{app:worked-reaction-diffusion}

Consider the one-dimensional equation
\begin{equation}
	\label{eq:app_rd_true}
	\partial_t u
	=
	\kappa u
	+
	D\partial_{xx}u ,
	\qquad x\in(0,1),\quad t\in(0,T),
\end{equation}
with known boundary and initial data. Suppose observations of \(u\) are available on sampling points
\[
z_i=(x_i,t_i),\qquad i=1,\ldots,n .
\]
Let the candidate library be
\[
\Gamma=\{u,\partial_{xx}u,u^2,\partial_xu\}.
\]
The anchored regression uses \(\partial_tu\) as the left-hand side:
\begin{equation}
	\label{eq:app_anchor_rd}
	y_i=\partial_tu(z_i),
	\qquad
	A_{i1}=u(z_i),
	\qquad
	A_{i2}=\partial_{xx}u(z_i),
	\qquad
	A_{i3}=u(z_i)^2,
	\qquad
	A_{i4}=\partial_xu(z_i).
\end{equation}
The coefficient vector is
\[
\beta^\star=(\kappa,D,0,0)^\top .
\]
A sparse screen solves
\begin{equation}
	\label{eq:app_lasso_rd}
	\widehat{\beta}
	\in
	\arg\min_{\beta\in\mathbb{R}^4}
	\left\{
	\frac{1}{2n}\|A\beta-y\|_2^2+\lambda\|\beta\|_1
	\right\}.
\end{equation}
If the design satisfies the restricted-eigenvalue condition in Assumption~\ref{ass:RE}, and the noise obeys the bound in Theorem~\ref{thm:lasso_screening_error}, then \(\widehat{\beta}\) is close to \(\beta^\star\). A thresholded support
\[
\widehat{S}_{\tau}
=
\{j:|\widehat{\beta}_j|>\tau\}
\]
is then used as a screened candidate support.

The counterfactual step does not ask whether a coefficient is merely nonzero. It asks whether deleting the term changes the induced trajectory. For \(j\in\widehat{S}_{\tau}\), define the intervened equation
\begin{equation}
	\label{eq:app_rd_cf}
	\partial_t u_j^{\cf}
	=
	\sum_{k\in\widehat{S}_{\tau},\,k\neq j}
	\widehat{\beta}_k\mathcal{T}_k[u_j^{\cf}]
\end{equation}
with the same initial and boundary data. The deletion score is
\[
\widehat{\delta}_j
=
\|u-u_j^{\cf}\|_{L^2((0,1)\times(0,T))}.
\]
The final relevance set is
\[
\widehat{S}_{\cf}
=
\{j\in\widehat{S}_{\tau}:\widehat{\delta}_j\ge\varepsilon\}.
\]
Thus coefficient recovery and counterfactual relevance are distinct. The sparse regression step estimates a candidate PDE. The deletion step tests which retained terms are functionally needed for the observed trajectory.

\subsection{Aliasing in a single-mode experiment}
\label{app:worked-aliasing}

The same model can become non-identifiable if the experiment excites only one Laplacian eigenmode. Let
\[
u(x,t)=a(t)\sin(\pi x).
\]
Then
\[
\partial_{xx}u=-\pi^2u.
\]
Substituting this identity into \eqref{eq:app_rd_true} gives
\[
\partial_tu
=
(\kappa-\pi^2D)u .
\]
Therefore the experiment identifies only the combined coefficient
\[
r=\kappa-\pi^2D.
\]
There are infinitely many pairs \((\kappa,D)\) satisfying the same observed relation:
\[
\kappa-\pi^2D=r .
\]
The design matrix has collinear columns
\[
A_{\partial_{xx}u}=-\pi^2A_u .
\]
This is exactly the aliasing obstruction in Theorem~\ref{thm:aliasing_obstruction}. No sparse-regression method can recover \(\kappa\) and \(D\) separately from this experiment alone. A counterfactual deletion test may still determine whether a fitted term affects the induced trajectory under the fitted model, but it cannot turn an aliased experiment into an identifiable coefficient-recovery problem.

\subsection{Breaking aliasing with two modes}
\label{app:worked-two-mode}

Now suppose the experiment excites two Laplacian eigenmodes:
\[
u(x,t)
=
a_1(t)\sin(\pi x)
+
a_2(t)\sin(2\pi x).
\]
For the first mode,
\[
\partial_{xx}\sin(\pi x)=-\pi^2\sin(\pi x),
\]
while for the second mode,
\[
\partial_{xx}\sin(2\pi x)=-4\pi^2\sin(2\pi x).
\]
If the modal growth rates are \(r_1\) and \(r_2\), then
\[
r_1=\kappa-\pi^2D,
\qquad
r_2=\kappa-4\pi^2D .
\]
Solving gives
\begin{equation}
	\label{eq:app_two_mode_recovery}
	D=\frac{r_1-r_2}{3\pi^2},
	\qquad
	\kappa=\frac{4r_1-r_2}{3}.
\end{equation}
This illustrates Theorem~\ref{thm:multi_experiment_separability}. Additional trajectories or modes improve identifiability only when they excite different operator directions. In matrix terms, the columns \(u\) and \(\partial_{xx}u\) are no longer perfectly collinear on the stacked experiment.

\subsection{Constraint-null pruning in a Maxwell-type system}
\label{app:worked-maxwell}

Let \(E\) be a divergence-free electric field satisfying
\[
\nabla\cdot E=0 .
\]
Consider an overcomplete library containing the candidate operator
\[
G(E)=\nabla(\nabla\cdot E).
\]
On the divergence-free constraint class,
\[
G(E)=0.
\]
Thus the coefficient of \(G\) is not identifiable from trajectories restricted to this class. This follows from Theorem~\ref{thm:constraint_null_nonidentifiability}. If a fitted residual includes
\[
\alpha_G\nabla(\nabla\cdot E),
\]
then changing \(\alpha_G\) does not change any residual evaluation on divergence-free data. The correct conclusion is not that \(\alpha_G=0\) has been estimated with high confidence. The correct conclusion is that the term is constraint-null and should be removed or grouped before interpreting coefficients.

The counterfactual deletion score also vanishes:
\[
\delta_G
=
\|E-E_G^{\cf}\|
\approx0
\]
up to numerical error. This is the operational meaning of constraint-aware library reduction.

\subsection{Observable-level relevance}
\label{app:worked-observable}

In some applications, full-state deviation is too strong. Suppose the scientific target is the terminal mass
\[
\mathcal{O}[u]
=
\int_\Omega u(x,T)\,dx .
\]
For deletion of \(\mathcal{T}_j\), the observable deletion score is
\[
\Delta_j^{\mathcal{O}}
=
|\mathcal{O}(u)-\mathcal{O}(u_j^{\cf})|.
\]
The adjoint approximation in Theorem~\ref{thm:observable_adjoint_relevance} gives the first-order score
\[
I_j^{\mathcal{O}}
=
|\langle\phi,\alpha_j\mathcal{T}_j[u]\rangle|,
\]
where
\[
L_u^*\phi=D\mathcal{O}(u).
\]
This score is useful for ranking candidate deletions when solving every counterfactual system is expensive. It should be interpreted as a first-order diagnostic, not as a replacement for finite counterfactual deletion when the intervention is large.

\subsection{Workflow summary}
\label{app:worked-workflow}

The complete workflow used throughout the paper is:

\begin{enumerate}[label=(\alph*)]
	\item Choose a candidate library \(\Gamma=\{\mathcal{T}_j\}_{j=1}^m\).
	\item Put the PDE relation in anchored or scale-normalized regression form.
	\item Build the operator-evaluation matrix \(A\).
	\item Report design diagnostics such as coherence and restricted-eigenvalue proxies.
	\item Use sparse regression to obtain a recall-oriented candidate support.
	\item De-bias coefficients on the selected support.
	\item Delete or perturb each selected operator.
	\item Compute state-level or observable-level counterfactual deviation.
	\item Report relevant, irrelevant, and unresolved terms at the chosen tolerance.
\end{enumerate}

This workflow explains how the theory is used to discover candidate PDE coefficients and to test whether those coefficients correspond to functional operator relevance.

\section{Additional Validation Diagnostics}
\label{app:validation-diagnostics}

This appendix reports the additional diagnostics used to support the main validation section. The appendix focuses on four issues: bootstrap coefficient stability, bootstrap selection paths, Lasso regularization-path behaviour, and full synthetic operator-level diagnostics. These results are not used to claim unconditional discovery of physical laws. They document the finite-sample behaviour of the screening and counterfactual-pruning pipeline under the reported experiments.

\subsection{Bootstrap coefficient and stability diagnostics}

\begin{figure*}[!t]
	\centering
	\begin{subfigure}{0.24\textwidth}
		\centering
		\includegraphics[width=\linewidth]{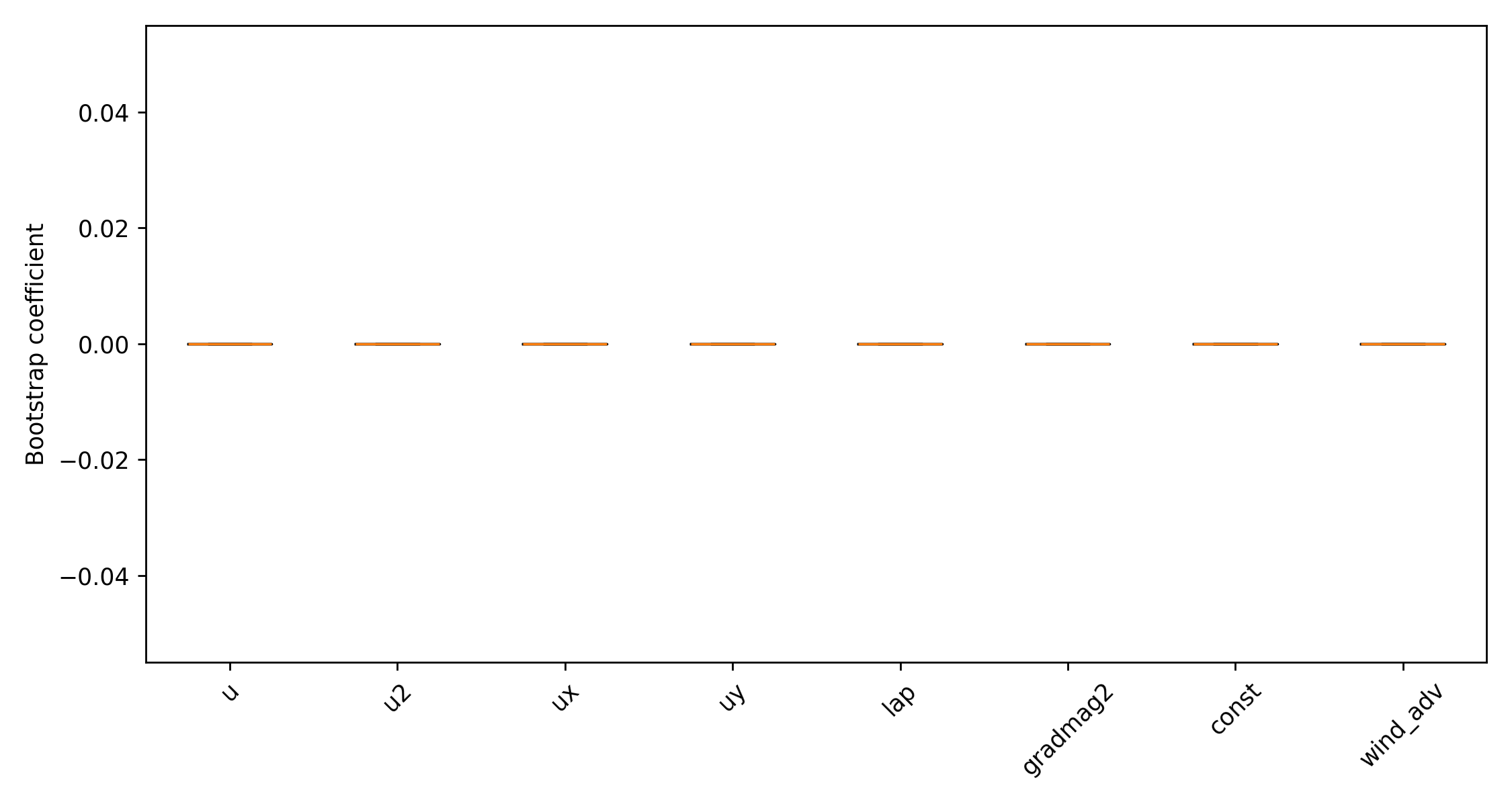}
		\caption{Atmospheric bootstrap coefficients.}
		\label{fig:boot_coef_era5}
	\end{subfigure}
	\hfill
	\begin{subfigure}{0.24\textwidth}
		\centering
		\includegraphics[width=\linewidth]{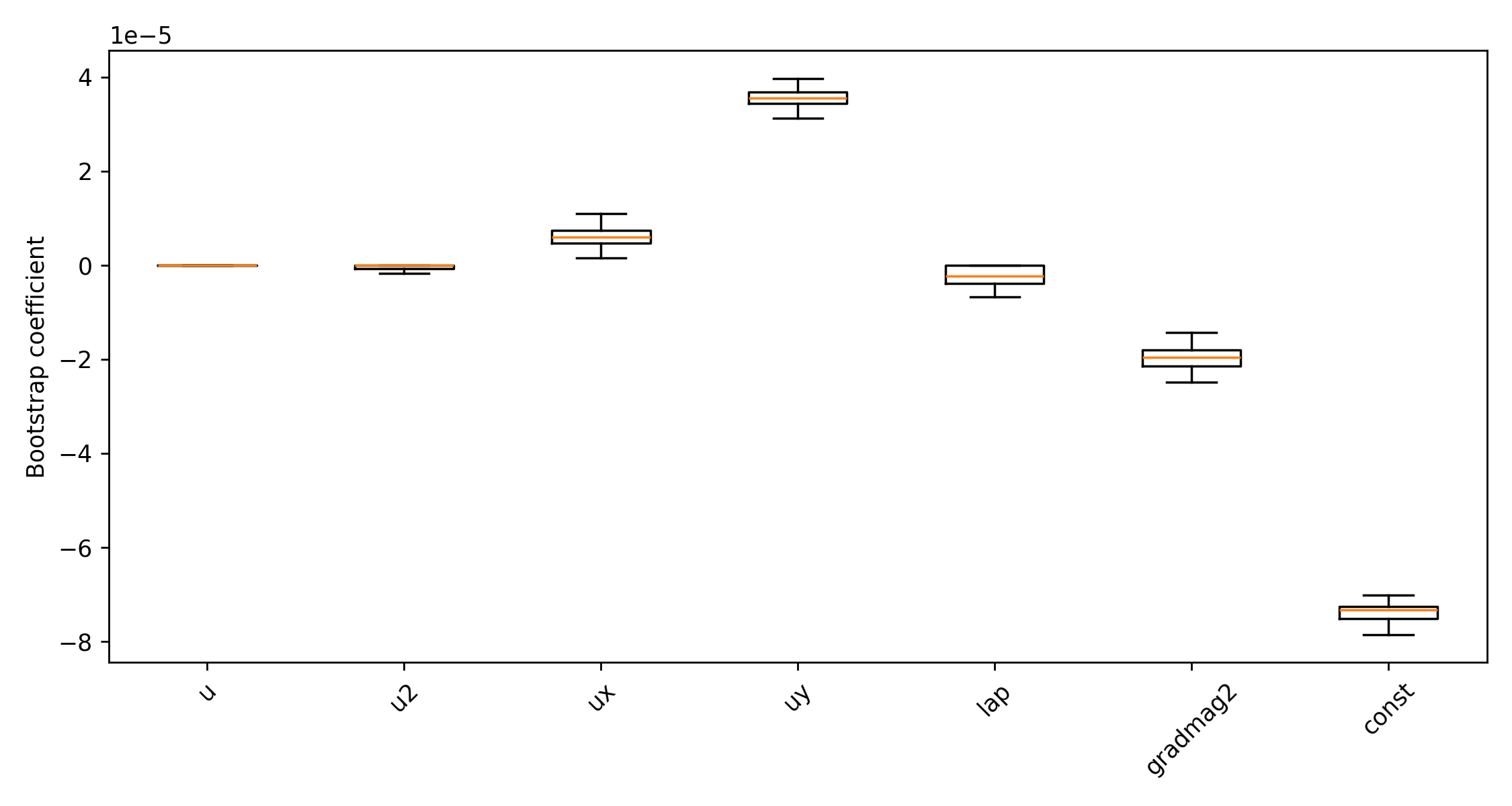}
		\caption{OISST bootstrap coefficients.}
		\label{fig:boot_coef_oisst}
	\end{subfigure}
	\hfill
	\begin{subfigure}{0.24\textwidth}
		\centering
		\includegraphics[width=\linewidth]{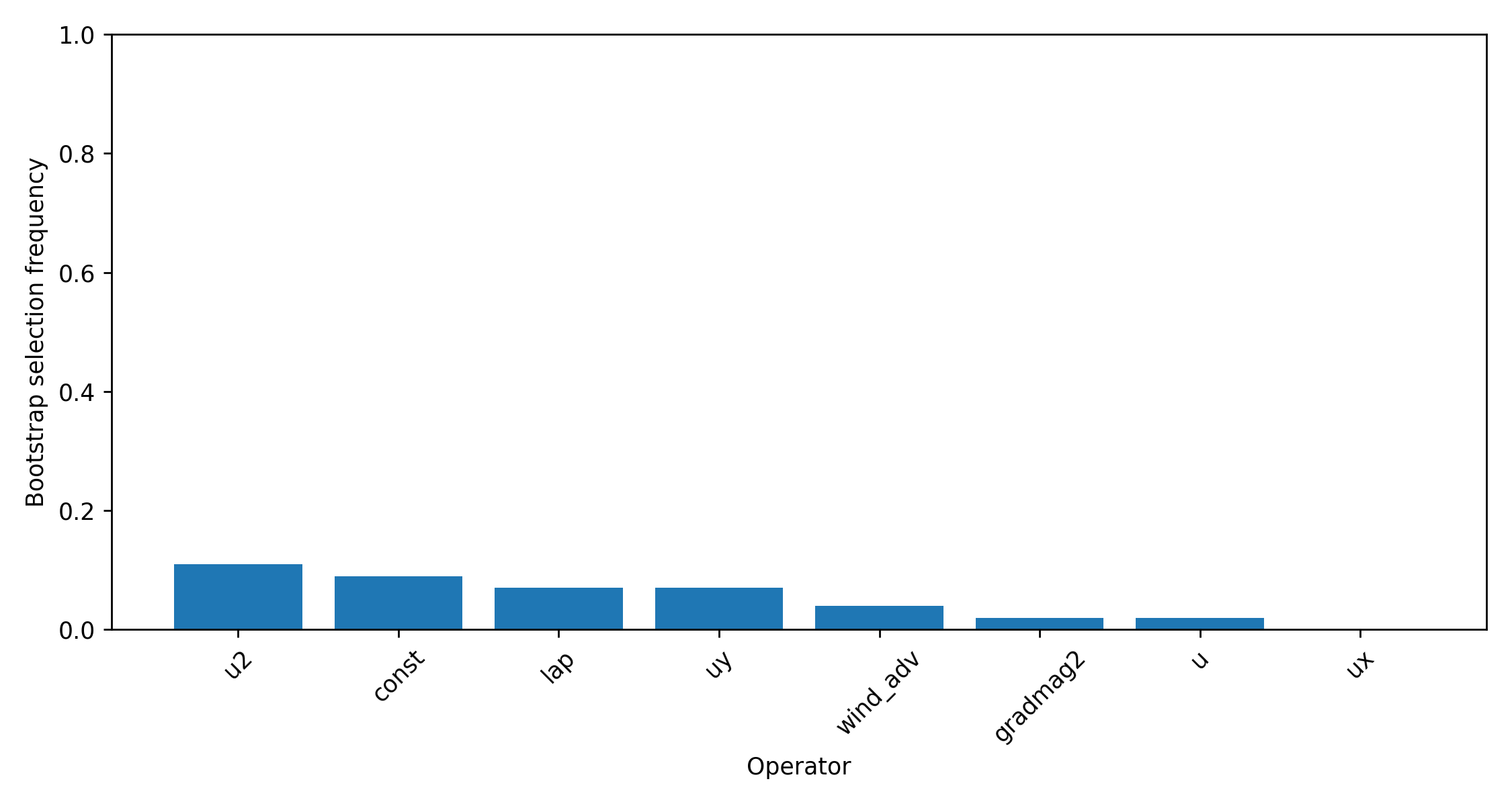}
		\caption{Atmospheric stability path.}
		\label{fig:stab_era5}
	\end{subfigure}
	\hfill
	\begin{subfigure}{0.24\textwidth}
		\centering
		\includegraphics[width=\linewidth]{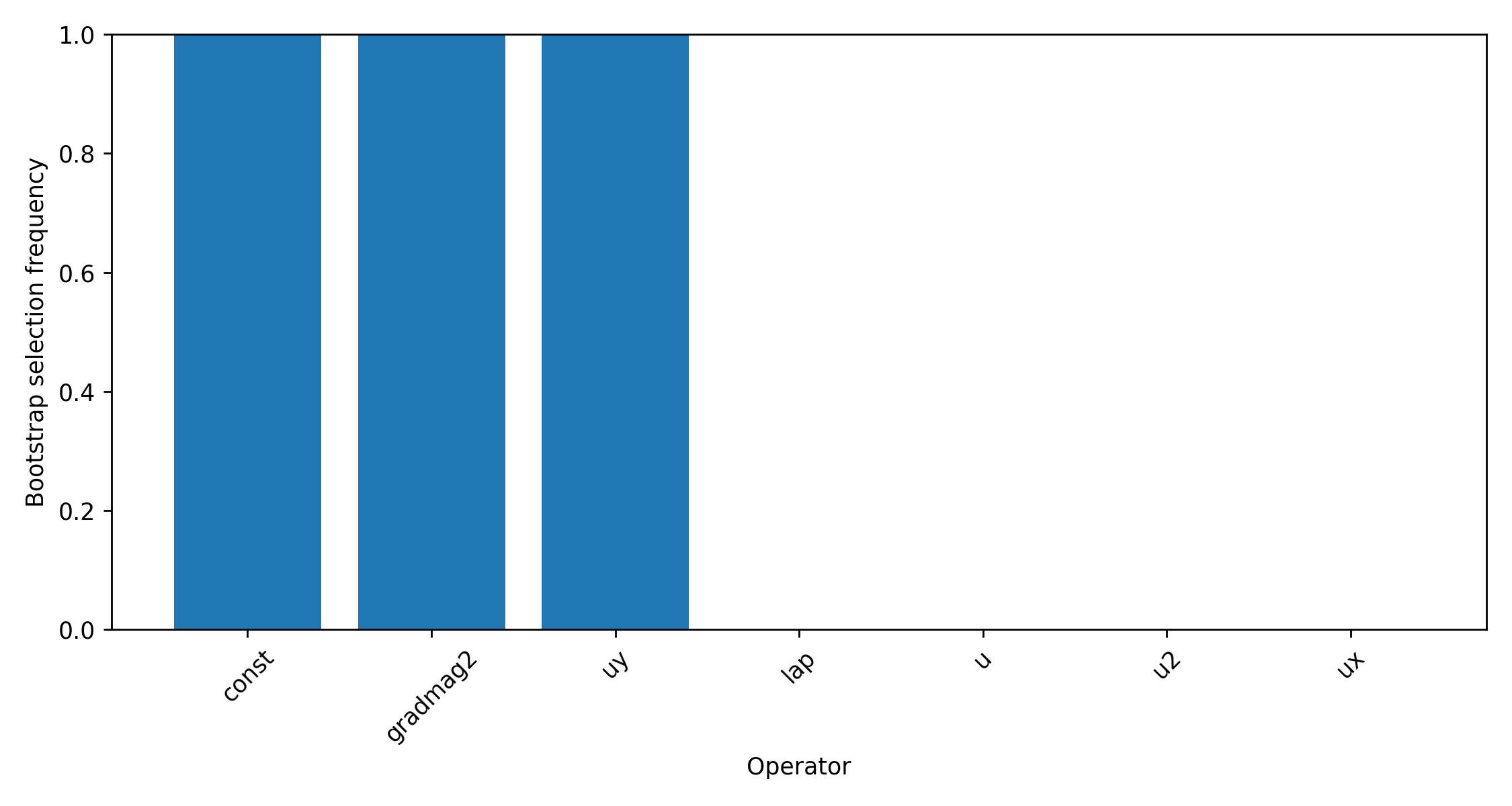}
		\caption{OISST stability path.}
		\label{fig:stab_oisst}
	\end{subfigure}
	\caption{Appendix bootstrap diagnostics for the real-data experiments. Panels~\ref{fig:boot_coef_era5} and~\ref{fig:boot_coef_oisst} show bootstrap coefficient distributions. Panels~\ref{fig:stab_era5} and~\ref{fig:stab_oisst} show bootstrap selection frequencies.}
	\label{fig:appendix_bootstrap_stability}
\end{figure*}

Figure~\ref{fig:appendix_bootstrap_stability} shows the real-data bootstrap coefficient distributions and selection frequencies. The atmospheric reanalysis case is unstable under bootstrap resampling, with no operator selected consistently. In contrast, the OISST case has stable bootstrap selection of the constant, \(|\nabla u|^2\), and \(u_y\) terms, each selected in all bootstrap replicates. This supports the main-text interpretation that the OISST case is the more informative real-data operator-surrogate experiment.

Table~\ref{tab:appendix_stability_paths} gives the bootstrap stability paths. In the atmospheric reanalysis case, the largest selection frequency is only \(0.11\), attained by \(u^2\). The constant term has frequency \(0.09\), while \(u_y\) and \(\Delta u\) have frequency \(0.07\). This confirms that the atmospheric run should be treated as unresolved. In the OISST case, the constant term, \(|\nabla u|^2\), and \(u_y\) have selection frequency \(1.00\), while \(u\), \(u^2\), \(u_x\), and \(\Delta u\) have selection frequency \(0.00\).

\begin{table*}[!t]
	\centering
	\caption{Bootstrap selection stability for the real-data experiments.}
	\label{tab:appendix_stability_paths}
	\renewcommand{\arraystretch}{1.15}
	\resizebox{\textwidth}{!}{
		\begin{tabular}{llccc}
			\toprule
			\textbf{Dataset}
			& \textbf{Operator}
			& \textbf{Selection frequency}
			& \textbf{Coefficient mean}
			& \textbf{Coefficient standard deviation}
			\\
			\midrule
			Atmospheric reanalysis
			& \(u^2\)
			& \(0.11\)
			& \(9.370\times10^{-7}\)
			& \(2.851\times10^{-6}\)
			\\
			Atmospheric reanalysis
			& constant
			& \(0.09\)
			& \(1.308\times10^{-6}\)
			& \(4.495\times10^{-6}\)
			\\
			Atmospheric reanalysis
			& \(u_y\)
			& \(0.07\)
			& \(-9.400\times10^{-7}\)
			& \(3.576\times10^{-6}\)
			\\
			Atmospheric reanalysis
			& \(\Delta u\)
			& \(0.07\)
			& \(-3.861\times10^{-7}\)
			& \(3.220\times10^{-6}\)
			\\
			Atmospheric reanalysis
			& wind advection
			& \(0.04\)
			& \(-7.871\times10^{-7}\)
			& \(4.100\times10^{-6}\)
			\\
			Atmospheric reanalysis
			& \(|\nabla u|^2\)
			& \(0.02\)
			& \(1.394\times10^{-7}\)
			& \(8.763\times10^{-7}\)
			\\
			Atmospheric reanalysis
			& \(u\)
			& \(0.02\)
			& \(1.969\times10^{-7}\)
			& \(1.448\times10^{-6}\)
			\\
			Atmospheric reanalysis
			& \(u_x\)
			& \(0.00\)
			& \(0.000\)
			& \(0.000\)
			\\
			\midrule
			OISST
			& constant
			& \(1.00\)
			& \(-7.372\times10^{-5}\)
			& \(1.867\times10^{-6}\)
			\\
			OISST
			& \(|\nabla u|^2\)
			& \(1.00\)
			& \(-1.966\times10^{-5}\)
			& \(2.432\times10^{-6}\)
			\\
			OISST
			& \(u_y\)
			& \(1.00\)
			& \(3.551\times10^{-5}\)
			& \(1.950\times10^{-6}\)
			\\
			OISST
			& \(\Delta u\)
			& \(0.00\)
			& \(-2.334\times10^{-6}\)
			& \(1.933\times10^{-6}\)
			\\
			OISST
			& \(u\)
			& \(0.00\)
			& \(5.654\times10^{-7}\)
			& \(1.085\times10^{-6}\)
			\\
			OISST
			& \(u^2\)
			& \(0.00\)
			& \(-5.443\times10^{-7}\)
			& \(1.068\times10^{-6}\)
			\\
			OISST
			& \(u_x\)
			& \(0.00\)
			& \(6.020\times10^{-6}\)
			& \(1.943\times10^{-6}\)
			\\
			\bottomrule
	\end{tabular}}
\end{table*}

Table~\ref{tab:appendix_bootstrap_intervals} reports empirical \(95\%\) bootstrap coefficient intervals. The atmospheric intervals are either degenerate at zero or include zero, which agrees with the unresolved classification. For OISST, the constant, \(|\nabla u|^2\), and \(u_y\) terms have nonzero intervals away from zero. The \(u_x\) coefficient also has a positive bootstrap interval, but its counterfactual deletion score in the main text is below the relevance threshold. This is useful: it shows why coefficient stability and counterfactual relevance are not the same diagnostic.

\begin{table*}[!t]
	\centering
	\caption{Bootstrap coefficient intervals for real-data experiments. Intervals are empirical \(2.5\%\) and \(97.5\%\) quantiles over bootstrap replicates.}
	\label{tab:appendix_bootstrap_intervals}
	\renewcommand{\arraystretch}{1.15}
	\resizebox{\textwidth}{!}{
		\begin{tabular}{llcccc}
			\toprule
			\textbf{Dataset}
			& \textbf{Operator}
			& \textbf{Selection frequency}
			& \textbf{Coefficient mean}
			& \textbf{2.5\% quantile}
			& \textbf{97.5\% quantile}
			\\
			\midrule
			Atmospheric reanalysis
			& \(u^2\)
			& \(0.11\)
			& \(9.370\times10^{-7}\)
			& \(0.000\)
			& \(1.098\times10^{-5}\)
			\\
			Atmospheric reanalysis
			& constant
			& \(0.09\)
			& \(1.308\times10^{-6}\)
			& \(0.000\)
			& \(1.675\times10^{-5}\)
			\\
			Atmospheric reanalysis
			& \(u_y\)
			& \(0.07\)
			& \(-9.400\times10^{-7}\)
			& \(-1.278\times10^{-5}\)
			& \(0.000\)
			\\
			Atmospheric reanalysis
			& \(\Delta u\)
			& \(0.07\)
			& \(-3.861\times10^{-7}\)
			& \(-9.985\times10^{-6}\)
			& \(0.000\)
			\\
			Atmospheric reanalysis
			& wind advection
			& \(0.04\)
			& \(-7.871\times10^{-7}\)
			& \(-1.419\times10^{-5}\)
			& \(0.000\)
			\\
			Atmospheric reanalysis
			& \(|\nabla u|^2\)
			& \(0.02\)
			& \(1.394\times10^{-7}\)
			& \(0.000\)
			& \(8.928\times10^{-7}\)
			\\
			Atmospheric reanalysis
			& \(u\)
			& \(0.02\)
			& \(1.969\times10^{-7}\)
			& \(0.000\)
			& \(0.000\)
			\\
			Atmospheric reanalysis
			& \(u_x\)
			& \(0.00\)
			& \(0.000\)
			& \(0.000\)
			& \(0.000\)
			\\
			\midrule
			OISST
			& constant
			& \(1.00\)
			& \(-7.372\times10^{-5}\)
			& \(-7.723\times10^{-5}\)
			& \(-7.047\times10^{-5}\)
			\\
			OISST
			& \(|\nabla u|^2\)
			& \(1.00\)
			& \(-1.966\times10^{-5}\)
			& \(-2.416\times10^{-5}\)
			& \(-1.486\times10^{-5}\)
			\\
			OISST
			& \(u_y\)
			& \(1.00\)
			& \(3.551\times10^{-5}\)
			& \(3.154\times10^{-5}\)
			& \(3.960\times10^{-5}\)
			\\
			OISST
			& \(u_x\)
			& \(0.00\)
			& \(6.020\times10^{-6}\)
			& \(2.017\times10^{-6}\)
			& \(9.299\times10^{-6}\)
			\\
			OISST
			& \(\Delta u\)
			& \(0.00\)
			& \(-2.334\times10^{-6}\)
			& \(-5.736\times10^{-6}\)
			& \(0.000\)
			\\
			OISST
			& \(u\)
			& \(0.00\)
			& \(5.654\times10^{-7}\)
			& \(0.000\)
			& \(3.408\times10^{-6}\)
			\\
			OISST
			& \(u^2\)
			& \(0.00\)
			& \(-5.443\times10^{-7}\)
			& \(-3.691\times10^{-6}\)
			& \(0.000\)
			\\
			\bottomrule
	\end{tabular}}
\end{table*}

\subsection{Lasso regularization-path diagnostics}

Table~\ref{tab:appendix_lasso_path} reports the BIC-selected Lasso model for the two real-data cases. The atmospheric case selects the zero model, with \(\alpha=5.621\times10^{-8}\), \(k=0\), residual sum of squares \(6.366\times10^{-9}\), and BIC \(-4176.771\). The OISST case selects a five-term model, with \(\alpha=8.381\times10^{-12}\), residual sum of squares \(3.001\times10^{-7}\), and BIC \(-6.710544\times10^{6}\). These results match the main-text interpretation: the atmospheric case gives no stable finite-support operator surrogate, whereas the OISST case gives a nontrivial sparse screen that is then pruned by counterfactual deletion.

\begin{table}[!t]
	\centering
	\caption{BIC-selected Lasso models for the real-data cases.}
	\label{tab:appendix_lasso_path}
	\renewcommand{\arraystretch}{1.15}
	\begin{tabular}{lcccc}
		\toprule
		\textbf{Dataset}
		& \(\boldsymbol{\alpha}\)
		& \textbf{RSS}
		& \(\boldsymbol{k}\)
		& \textbf{BIC}
		\\
		\midrule
		Atmospheric reanalysis
		& \(5.621\times10^{-8}\)
		& \(6.366\times10^{-9}\)
		& \(0\)
		& \(-4176.771\)
		\\
		OISST
		& \(8.381\times10^{-12}\)
		& \(3.001\times10^{-7}\)
		& \(5\)
		& \(-6.710544\times10^{6}\)
		\\
		\bottomrule
	\end{tabular}
\end{table}

\subsection{Full synthetic operator diagnostics}

Table~\ref{tab:appendix_synthetic_full} gives the full synthetic operator-level diagnostics. The table includes the selected rate, true-support indicator, mean coefficient, residual contribution index, mean deletion score, and decision frequencies. These values support the more compact main-text table.

\begin{sidewaystable}[p]
	\centering
	\scriptsize
	\caption{Full synthetic operator-level diagnostics over ten seeds.}
	\label{tab:appendix_synthetic_full}
	\renewcommand{\arraystretch}{1.08}
	\setlength{\tabcolsep}{4pt}
	\resizebox{\textheight}{!}{%
		\begin{tabular}{llcccccc}
			\toprule
			\textbf{Experiment}
			& \textbf{Operator}
			& \textbf{True active}
			& \textbf{Selection rate}
			& \textbf{Mean coefficient}
			& \textbf{Mean RCI}
			& \textbf{Mean deletion}
			& \textbf{Relevant rate}
			\\
			\midrule
			
			Residual--counterfactual gap
			& \(u\)
			& yes
			& \(1.00\)
			& \(6.012\times10^{-1}\)
			& \(1.002\)
			& \(6.310\times10^{-3}\)
			& \(1.00\)
			\\
			Residual--counterfactual gap
			& \(u^2\)
			& no
			& \(0.00\)
			& \(0.000\)
			& \(0.000\)
			& \(0.000\)
			& \(0.00\)
			\\
			Residual--counterfactual gap
			& \(u_x\)
			& no
			& \(0.00\)
			& \(0.000\)
			& \(0.000\)
			& \(0.000\)
			& \(0.00\)
			\\
			Residual--counterfactual gap
			& \(u_y\)
			& no
			& \(0.00\)
			& \(0.000\)
			& \(0.000\)
			& \(0.000\)
			& \(0.00\)
			\\
			Residual--counterfactual gap
			& \(\Delta u\)
			& no
			& \(0.00\)
			& \(3.762\times10^{-5}\)
			& \(4.764\times10^{-3}\)
			& \(3.000\times10^{-5}\)
			& \(0.80\)
			\\
			Residual--counterfactual gap
			& \(|\nabla u|^2\)
			& no
			& \(0.00\)
			& \(0.000\)
			& \(0.000\)
			& \(0.000\)
			& \(0.00\)
			\\
			Residual--counterfactual gap
			& constant
			& no
			& \(0.00\)
			& \(0.000\)
			& \(0.000\)
			& \(0.000\)
			& \(0.00\)
			\\
			\addlinespace
			
			Single-mode aliasing
			& \(u\)
			& yes
			& \(1.00\)
			& \(6.000\times10^{-1}\)
			& \(1.000\)
			& \(5.634\times10^{-3}\)
			& \(1.00\)
			\\
			Single-mode aliasing
			& \(u^2\)
			& no
			& \(0.00\)
			& \(0.000\)
			& \(0.000\)
			& \(0.000\)
			& \(0.00\)
			\\
			Single-mode aliasing
			& \(u_x\)
			& no
			& \(0.00\)
			& \(0.000\)
			& \(0.000\)
			& \(0.000\)
			& \(0.00\)
			\\
			Single-mode aliasing
			& \(u_y\)
			& no
			& \(0.00\)
			& \(0.000\)
			& \(0.000\)
			& \(0.000\)
			& \(0.00\)
			\\
			Single-mode aliasing
			& \(\Delta u\)
			& no
			& \(0.00\)
			& \(0.000\)
			& \(0.000\)
			& \(0.000\)
			& \(0.00\)
			\\
			Single-mode aliasing
			& \(|\nabla u|^2\)
			& no
			& \(0.00\)
			& \(0.000\)
			& \(0.000\)
			& \(0.000\)
			& \(0.00\)
			\\
			Single-mode aliasing
			& constant
			& no
			& \(0.00\)
			& \(0.000\)
			& \(0.000\)
			& \(0.000\)
			& \(0.00\)
			\\
			\addlinespace
			
			Multi-trajectory separability
			& \(u\)
			& yes
			& \(1.00\)
			& \(4.892\times10^{-1}\)
			& \(9.554\times10^{-1}\)
			& \(5.075\times10^{-3}\)
			& \(1.00\)
			\\
			Multi-trajectory separability
			& \(u^2\)
			& no
			& \(1.00\)
			& \(2.424\times10^{-2}\)
			& \(5.796\times10^{-2}\)
			& \(3.079\times10^{-4}\)
			& \(1.00\)
			\\
			Multi-trajectory separability
			& \(u_x\)
			& no
			& \(0.00\)
			& \(-3.817\times10^{-3}\)
			& \(2.692\times10^{-2}\)
			& \(1.430\times10^{-4}\)
			& \(0.00\)
			\\
			Multi-trajectory separability
			& \(u_y\)
			& no
			& \(0.00\)
			& \(0.000\)
			& \(0.000\)
			& \(0.000\)
			& \(0.00\)
			\\
			Multi-trajectory separability
			& \(\Delta u\)
			& yes
			& \(0.00\)
			& \(8.816\times10^{-4}\)
			& \(8.144\times10^{-2}\)
			& \(4.326\times10^{-4}\)
			& \(1.00\)
			\\
			Multi-trajectory separability
			& \(|\nabla u|^2\)
			& no
			& \(0.00\)
			& \(-2.284\times10^{-4}\)
			& \(1.252\times10^{-2}\)
			& \(6.653\times10^{-5}\)
			& \(0.00\)
			\\
			Multi-trajectory separability
			& constant
			& no
			& \(1.00\)
			& \(2.302\times10^{-2}\)
			& \(5.486\times10^{-2}\)
			& \(2.914\times10^{-4}\)
			& \(0.00\)
			\\
			\addlinespace
			
			Advection--diffusion
			& \(u\)
			& no
			& \(1.00\)
			& \(-1.503\times10^{-1}\)
			& \(2.047\times10^{-1}\)
			& \(9.239\times10^{-4}\)
			& \(1.00\)
			\\
			Advection--diffusion
			& \(u^2\)
			& no
			& \(1.00\)
			& \(-2.222\times10^{-2}\)
			& \(3.410\times10^{-2}\)
			& \(1.539\times10^{-4}\)
			& \(0.00\)
			\\
			Advection--diffusion
			& \(u_x\)
			& yes
			& \(1.00\)
			& \(-1.586\times10^{-1}\)
			& \(8.948\times10^{-1}\)
			& \(4.038\times10^{-3}\)
			& \(1.00\)
			\\
			Advection--diffusion
			& \(u_y\)
			& yes
			& \(1.00\)
			& \(8.032\times10^{-2}\)
			& \(4.242\times10^{-1}\)
			& \(1.914\times10^{-3}\)
			& \(1.00\)
			\\
			Advection--diffusion
			& \(\Delta u\)
			& yes
			& \(0.00\)
			& \(2.943\times10^{-4}\)
			& \(2.122\times10^{-2}\)
			& \(9.574\times10^{-5}\)
			& \(0.00\)
			\\
			Advection--diffusion
			& \(|\nabla u|^2\)
			& no
			& \(0.00\)
			& \(6.416\times10^{-4}\)
			& \(2.295\times10^{-2}\)
			& \(1.036\times10^{-4}\)
			& \(0.00\)
			\\
			Advection--diffusion
			& constant
			& no
			& \(1.00\)
			& \(4.326\times10^{-2}\)
			& \(9.707\times10^{-2}\)
			& \(4.380\times10^{-4}\)
			& \(0.00\)
			\\
			\addlinespace
			
			Constraint-null
			& \(\nabla(\nabla\cdot E)\)
			& no
			& \(0.00\)
			& \(1.000\)
			& \(1.329\times10^{-13}\)
			& \(3.012\times10^{-15}\)
			& \(0.00\)
			\\
			
			\bottomrule
		\end{tabular}%
	}
\end{sidewaystable}

The appendix diagnostics reinforce the main conclusions without changing their scope. The atmospheric reanalysis experiment has weak and unstable sparse support, so it remains an unresolved real-data case. The OISST experiment has stable sparse selection of three terms, but counterfactual deletion in the main text retains only the terms with deletion scores above the stated threshold. The synthetic experiments expose both strengths and limitations of the current finite-sample implementation: dominant active terms are recovered reliably, while weak active terms such as diffusion in the advection--diffusion and multi-trajectory experiments may require recall-oriented screening or stronger excitation before counterfactual pruning.

\bibliographystyle{unsrtnat}
\bibliography{file}
\end{document}